\documentclass[10pt,twocolumn,letterpaper]{article}

\usepackage{cvpr}              

\usepackage[dvipsnames]{xcolor}

\definecolor{cvprblue}{rgb}{0.21,0.49,0.74}
\usepackage[pagebackref,breaklinks,colorlinks,citecolor=cvprblue]{hyperref}
\usepackage{multirow}
\usepackage{xcolor}
\usepackage{comment}
\usepackage{pifont}
\usepackage{multirow}
\usepackage{float}
\usepackage{makecell}
\usepackage{amsmath}
\usepackage{algpseudocode}
\usepackage{algorithm}
\usepackage{lettrine}

\DeclareMathOperator*{\argmin}{argmin} 
\newcommand{\xmark}{\ding{55}}%

\usepackage[
singlelinecheck=false 
]{caption}
\def\paperID{3015} 
\def\confName{CVPR}
\def\confYear{2024}

\title{You Only Need One Step: Fast Super-Resolution with Stable Diffusion via Scale Distillation}

\author{Mehdi Noroozi, Isma Hadji, Brais Martinez, Adrian Bulat, Georgios Tzimiropoulos\\
Samsung AI Cambridge\\
{\tt\small m.noroozi@samsung.com}
}

\begin{document}
\maketitle

\begin{abstract}
In this paper, we introduce YONOS-SR, a novel stable diffusion-based approach for image super-resolution that yields state-of-the-art results using only a single DDIM step. We propose a novel scale distillation approach to train our SR model. Instead of directly training our SR model on the scale factor of interest, we start by training a teacher model on a smaller magnification scale, thereby making the SR problem simpler for the teacher.
We then train a student model for a higher magnification scale, using the predictions of the teacher as a target during the training. This process is repeated iteratively until we reach the target scale factor of the final model. The rationale behind our scale distillation is that the teacher aids the student diffusion model training by i) providing a target adapted to the current noise level rather than using the same target coming from ground truth data for all noise levels and ii) providing an accurate target as the teacher has a simpler task to solve. We empirically show that the distilled model significantly outperforms the model trained for high scales directly, specifically with few steps during inference. 
Having a strong diffusion model that requires only one step allows us to freeze the U-Net and fine-tune the decoder on top of it. We show that the combination of spatially distilled U-Net and fine-tuned decoder outperforms state-of-the-art methods requiring 200 steps with only one single step.

\end{abstract}
  
\section{Introduction}
\label{sec:intro}
\begin{figure*}[ht]
\begin{minipage}{0.33\textwidth}
  \begin{minipage}{0.49\textwidth}
      \begin{figure}[H]
      \centering
           \footnotesize{$\times 4$}
      \end{figure}
  \end{minipage} 
  \begin{minipage}{0.49\textwidth}
      \begin{figure}[H]
      \centering
        \footnotesize{$\times 8$}
      \end{figure}
  \end{minipage}
  \begin{minipage}{0.49\textwidth}
      \begin{figure}[H]
          \centering
          \includegraphics[width=\linewidth]{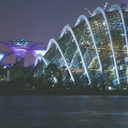}
      \end{figure}
  \end{minipage} 
    \begin{minipage}{0.49\textwidth}
      \begin{figure}[H]
          \centering
          \includegraphics[width=\linewidth]{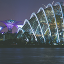}
      \end{figure}
  \end{minipage}
  \\
  \begin{minipage}{1\textwidth}
     \centering
      \footnotesize{~}
      \\
      \footnotesize{LR}
  \end{minipage}
  \\
  \begin{minipage}{0.49\textwidth}
      \begin{figure}[H]
          \centering
          \includegraphics[width=\linewidth]{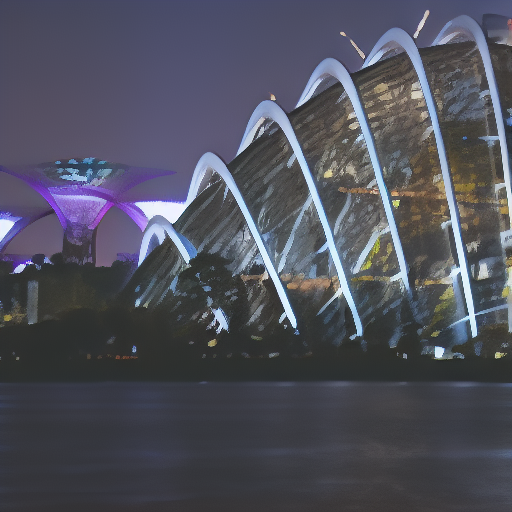}
      \end{figure}
  \end{minipage} 
    \begin{minipage}{0.49\textwidth}
      \begin{figure}[H]
          \centering
          \includegraphics[width=\linewidth]{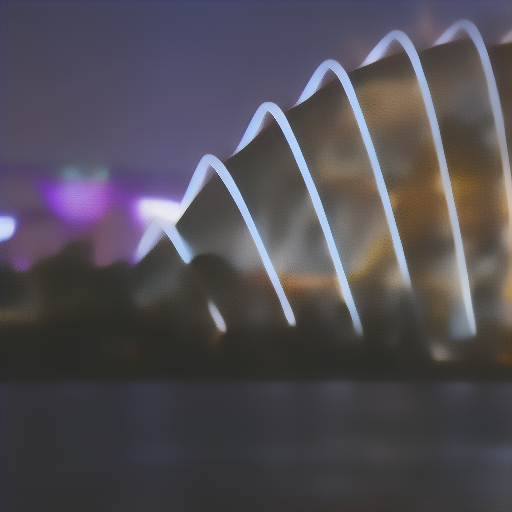}
      \end{figure}
  \end{minipage}
  \\
  \begin{minipage}{1\textwidth}
     \centering
     \footnotesize{~}
      \\
      \footnotesize{1-step SD-SR}
  \end{minipage}
  \\
  \begin{minipage}{0.49\textwidth}
      \begin{figure}[H]
          \centering
          \includegraphics[width=\linewidth]{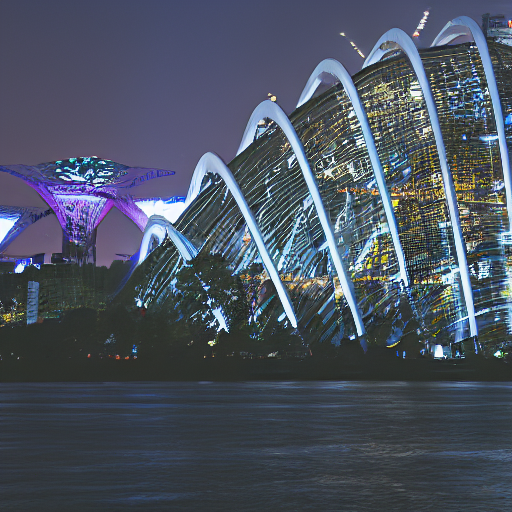}
      \end{figure}
  \end{minipage} 
    \begin{minipage}{0.49\textwidth}
      \begin{figure}[H]
          \centering
          \includegraphics[width=\linewidth]{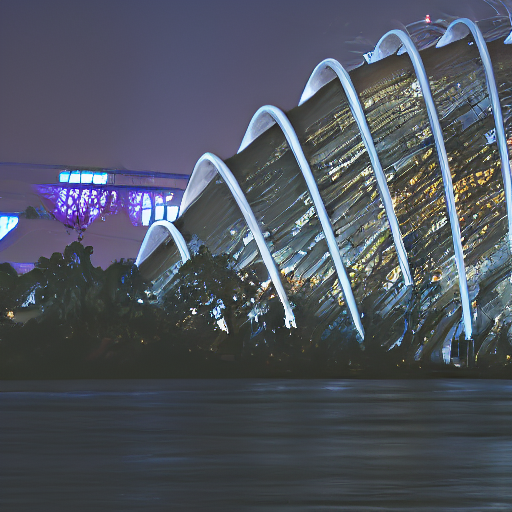}
      \end{figure}
  \end{minipage}
  \\
  \begin{minipage}{1\textwidth}
     \centering
      \footnotesize{~}
      \\
      \footnotesize{200-step SD-SR}
  \end{minipage}
  \\
  \begin{minipage}{0.49\textwidth}
      \begin{figure}[H]
          \centering
          \includegraphics[width=\linewidth]{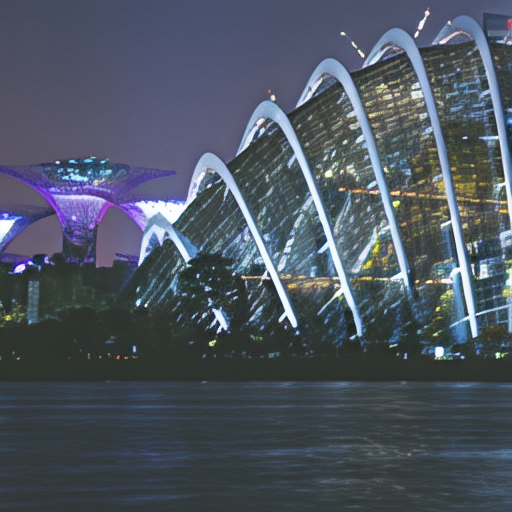}
      \end{figure}
  \end{minipage} 
    \begin{minipage}{0.49\textwidth}
      \begin{figure}[H]
          \centering
          \includegraphics[width=\linewidth]{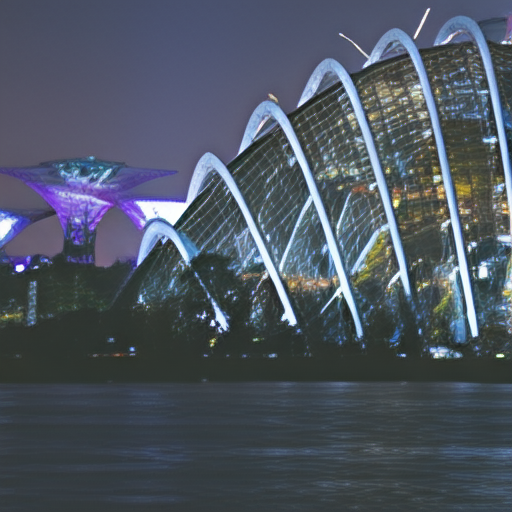}
      \end{figure}
  \end{minipage}
  \\
  \begin{minipage}{1\textwidth}
     \centering
      \footnotesize{~}
      \\
      \footnotesize{1-step YONOS-SR}
  \end{minipage}
  \\
   \begin{minipage}{1\textwidth}
      \begin{figure}[H]
          \centering
          \includegraphics[width=.55\linewidth]{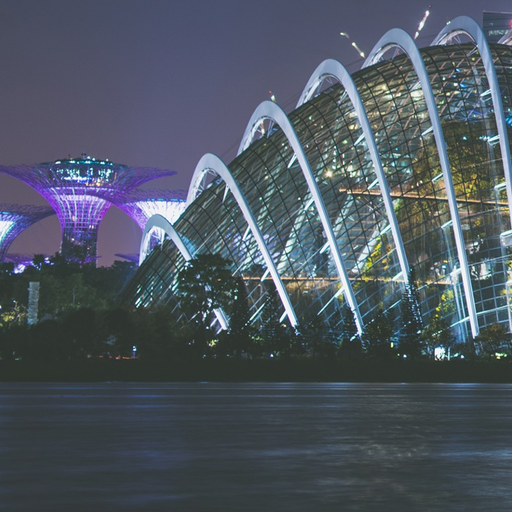}
      \end{figure}
  \end{minipage}
  \\
  \begin{minipage}{1\textwidth}
     \centering
           \footnotesize{~}
      \\
      \footnotesize{GT}
  \end{minipage}
\end{minipage}
\begin{minipage}{0.01\textwidth}
    \centering
    \textbf{
    \rotatebox{90}{ - - - - - - - - - - - - - - - - - - - - - - - - - - - - - - - - - - - - - - - - - - - - - - - - - - - - - - - - - - - - - - - - - - - - - - - - - - - - - - - - - - - - - - - -~~~~~~~~}}
\end{minipage} 
\begin{minipage}{0.33\textwidth}
  \begin{minipage}{0.49\textwidth}
      \begin{figure}[H]
      \centering
           \footnotesize{$\times 4$}
      \end{figure}
  \end{minipage} 
  \begin{minipage}{0.49\textwidth}
      \begin{figure}[H]
      \centering
        \footnotesize{$\times 8$}
      \end{figure}
  \end{minipage}
  \\
  \begin{minipage}{0.49\textwidth}
      \begin{figure}[H]
          \centering
          \includegraphics[width=\linewidth]{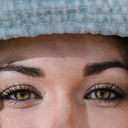}
      \end{figure}
  \end{minipage} 
    \begin{minipage}{0.49\textwidth}
      \begin{figure}[H]
          \centering
          \includegraphics[width=\linewidth]{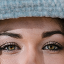}
      \end{figure}
  \end{minipage}
  \\
  \begin{minipage}{1\textwidth}
     \centering
      \footnotesize{~}
      \\
      \footnotesize{LR}
  \end{minipage}
  \\
  \begin{minipage}{0.49\textwidth}
      \begin{figure}[H]
          \centering
          \includegraphics[width=\linewidth]{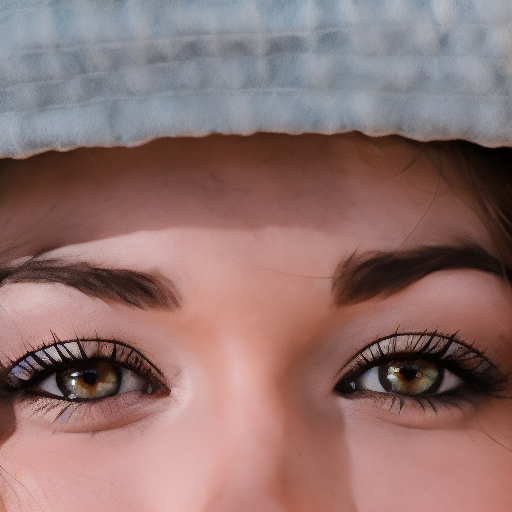}
      \end{figure}
  \end{minipage} 
    \begin{minipage}{0.49\textwidth}
      \begin{figure}[H]
          \centering
          \includegraphics[width=\linewidth]{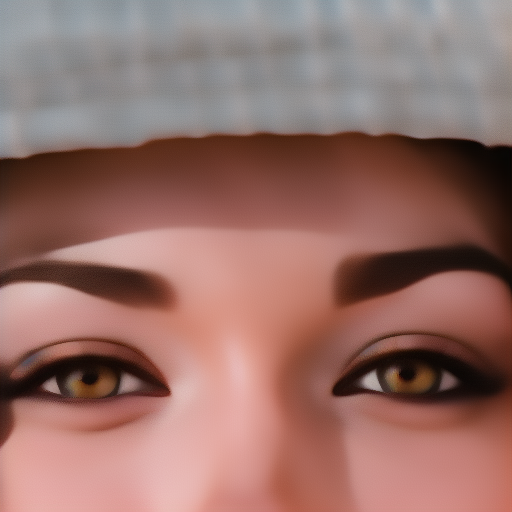}
      \end{figure}
  \end{minipage}
  \\
  \begin{minipage}{1\textwidth}
     \centering
      \footnotesize{~}
      \\
      \footnotesize{1-step SD-SR}
  \end{minipage}
  \\
  \begin{minipage}{0.49\textwidth}
      \begin{figure}[H]
          \centering
          \includegraphics[width=\linewidth]{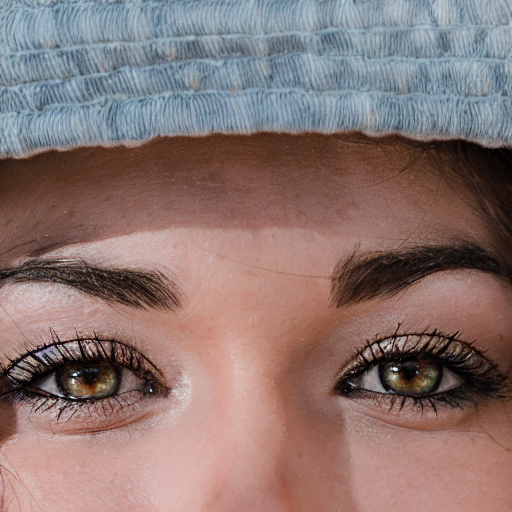}
      \end{figure}
  \end{minipage} 
    \begin{minipage}{0.49\textwidth}
      \begin{figure}[H]
          \centering
          \includegraphics[width=\linewidth]{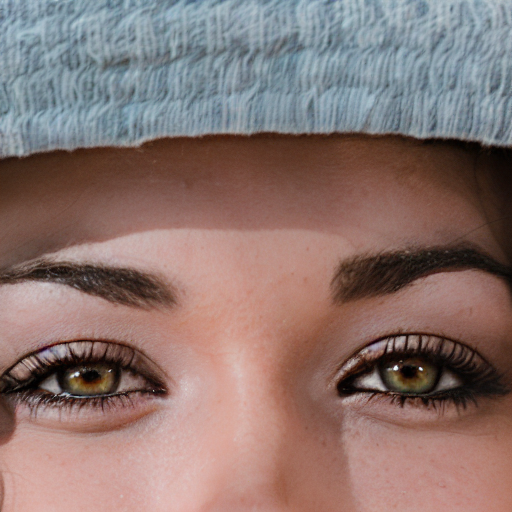}
      \end{figure}
  \end{minipage}
  \\
  \begin{minipage}{1\textwidth}
     \centering
      \footnotesize{~}
      \\
      \footnotesize{200-step SD-SR}
  \end{minipage}
  \\
  \begin{minipage}{0.49\textwidth}
      \begin{figure}[H]
          \centering
          \includegraphics[width=\linewidth]{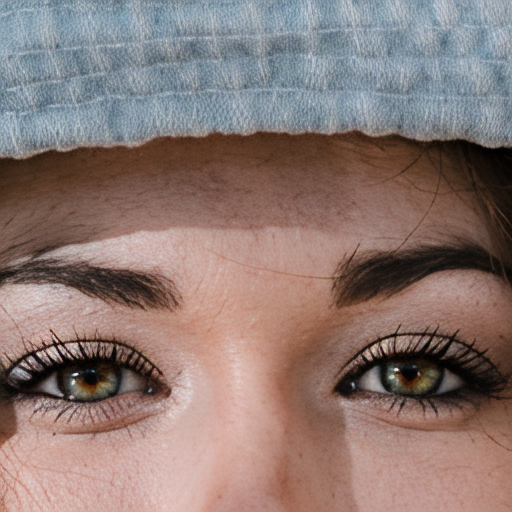}
      \end{figure}
  \end{minipage} 
    \begin{minipage}{0.49\textwidth}
      \begin{figure}[H]
          \centering
          \includegraphics[width=\linewidth]{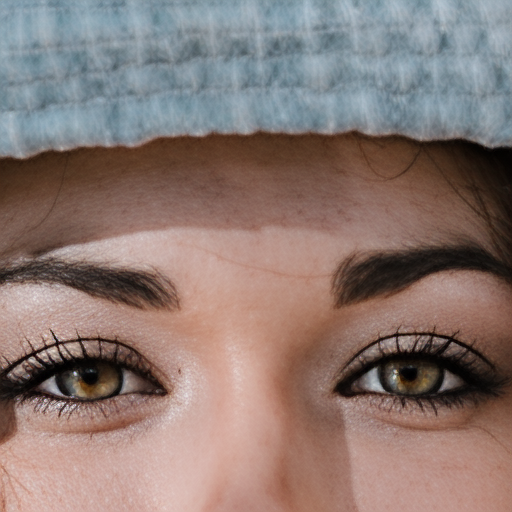}
      \end{figure}
  \end{minipage}
  \\
  \begin{minipage}{1\textwidth}
     \centering
      \footnotesize{~}
      \\
      \footnotesize{1-step YONOS-SR}
  \end{minipage}
  \\
   \begin{minipage}{1\textwidth}
      \begin{figure}[H]
          \centering
          \includegraphics[width=.55\linewidth]{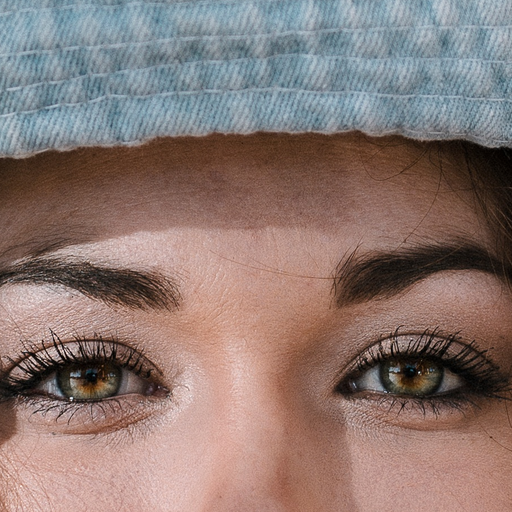}
      \end{figure}
  \end{minipage}
  \\
  \begin{minipage}{1\textwidth}
     \centering
    \footnotesize{~}
      \\
      \footnotesize{GT}
  \end{minipage}
\end{minipage}
\begin{minipage}{0.005\textwidth}
    \centering
    \textbf{
    \rotatebox{90}{ - - - - - - - - - - - - - - - - - - - - - - - - - - - - - - - - - - - - - - - - - - - - - - - - - - - - - - - - - - - - - - - - - - - - - - - - - - - - - - - - - - - - - - - -~~~~~~~~}}
\end{minipage} 
\begin{minipage}{0.33\textwidth}
  \begin{minipage}{0.49\textwidth}
      \begin{figure}[H]
      \centering
           \footnotesize{$\times 4$}
      \end{figure}
  \end{minipage} 
  \begin{minipage}{0.49\textwidth}
      \begin{figure}[H]
      \centering
        \footnotesize{$\times 8$}
      \end{figure}
  \end{minipage}
  \\
  \begin{minipage}{0.49\textwidth}
      \begin{figure}[H]
          \centering
          \includegraphics[width=\linewidth]{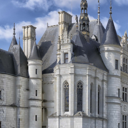}
      \end{figure}
  \end{minipage} 
    \begin{minipage}{0.49\textwidth}
      \begin{figure}[H]
          \centering
          \includegraphics[width=\linewidth]{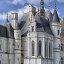}
      \end{figure}
  \end{minipage}
  \\
  \begin{minipage}{1\textwidth}
     \centering
     \footnotesize{~}
      \\
      \footnotesize{LR}
  \end{minipage}
  \\
  \begin{minipage}{0.49\textwidth}
      \begin{figure}[H]
          \centering
          \includegraphics[width=\linewidth]{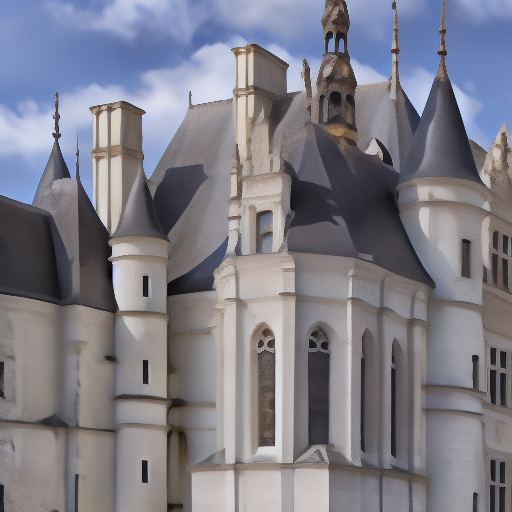}
      \end{figure}
  \end{minipage} 
    \begin{minipage}{0.49\textwidth}
      \begin{figure}[H]
          \centering
          \includegraphics[width=\linewidth]{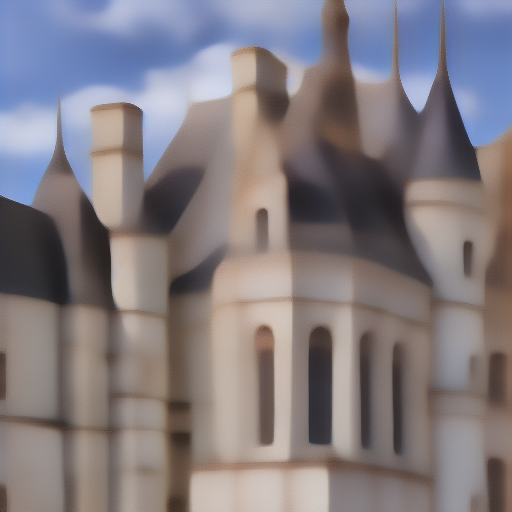}
      \end{figure}
  \end{minipage}
  \\
  \begin{minipage}{1\textwidth}
     \centering
       \footnotesize{~}
      \\
      \footnotesize{1-step SD-SR}
  \end{minipage}
  \\
  \begin{minipage}{0.49\textwidth}
      \begin{figure}[H]
          \centering
          \includegraphics[width=\linewidth]{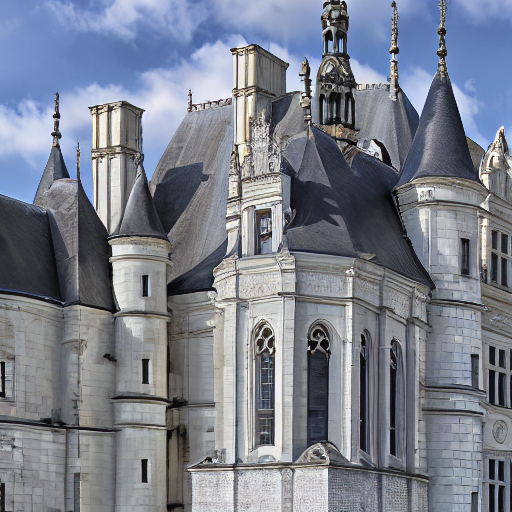}
      \end{figure}
  \end{minipage} 
    \begin{minipage}{0.49\textwidth}
      \begin{figure}[H]
          \centering
          \includegraphics[width=\linewidth]{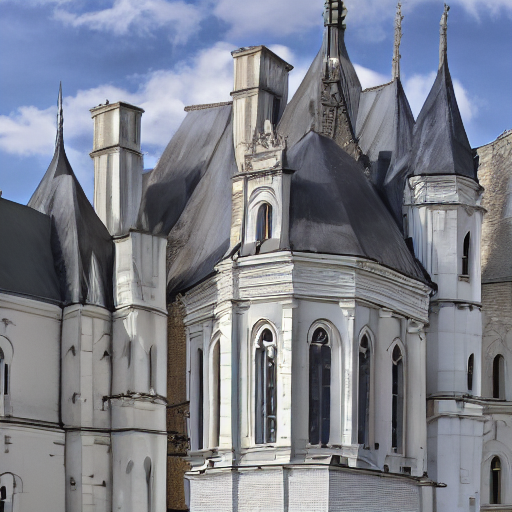}
      \end{figure}
  \end{minipage}
  \\
  \begin{minipage}{1\textwidth}
     \centering
      \footnotesize{~}
      \\
      \footnotesize{200-step SD-SR}
  \end{minipage}
  \\
  \begin{minipage}{0.49\textwidth}
      \begin{figure}[H]
          \centering
          \includegraphics[width=\linewidth]{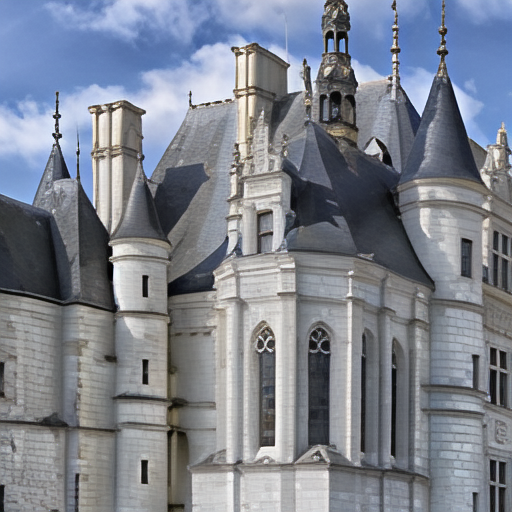}
      \end{figure}
  \end{minipage} 
    \begin{minipage}{0.49\textwidth}
      \begin{figure}[H]
          \centering
          \includegraphics[width=\linewidth]{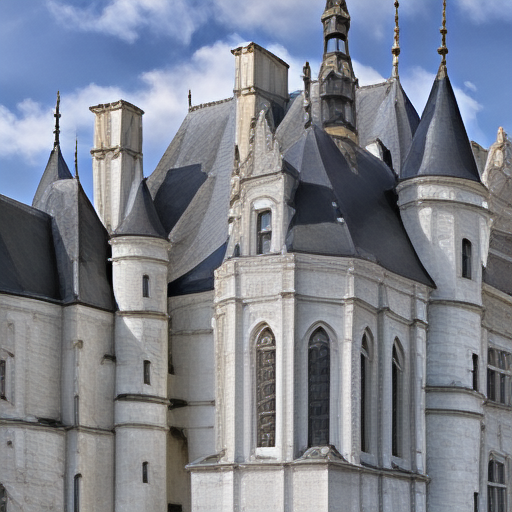}
      \end{figure}
  \end{minipage}
  \\
  \begin{minipage}{1\textwidth}
     \centering
      \footnotesize{~}
      \\
      \footnotesize{1-step YONOS-SR}
  \end{minipage}
  \\
   \begin{minipage}{1\textwidth}
      \begin{figure}[H]
          \centering
          \includegraphics[width=.55\linewidth]{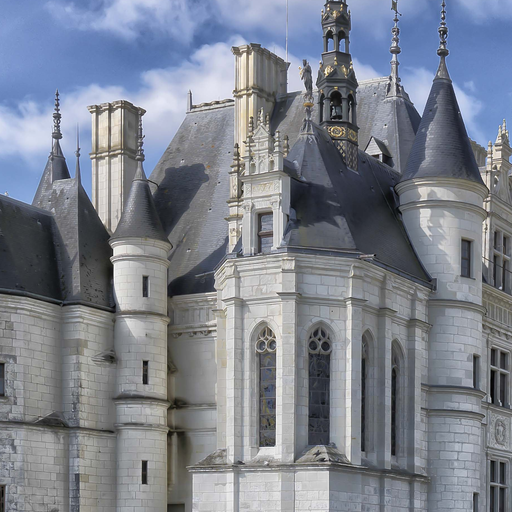}
      \end{figure}
  \end{minipage}
  \\
  \begin{minipage}{1\textwidth}
     \centering
      \footnotesize{~}
      \\
      \footnotesize{GT}
  \end{minipage}
\end{minipage}
\caption{Qualitative comparison for $\times 4$ and $\times 8$ magnifications.  Each column shows top to bottom LR input image, 1 and 200 step SD-SR,  and 1-step YONOS-SR(ours).  SD-SR represents the standard Stable Diffusion-based SR model, whereas YONOS-SR is our method trained using the same data and parameterization. The 1-step SD-SR method lacks quality in terms of detailed textures compared to 200-steps of the same model; see building texture in the first column and hairs in the middle column. In contrast, our proposed method outperforms 200-steps SD-SR with only one step specifically for $\times 8$ magnification where SD-SR fails to recover the details even with 200 steps. Samples are taken from DIV2K bicubic validation set. The images are best seen in a display and zoomed in.}
\label{fig-step-vs-quality}
\end{figure*}

Diffusion models have shown impressive performance in various image generation tasks~\cite{ldm_cvpr22,controlnet}, including image super-resolution (SR) \cite{stablesr_arx23, sr3,sr3+,p2l}.
However, the large number of sequential denoising passes required by the sampling strategy results in extreme computational cost, even for stable diffusion-based models (SD) that operate in the latent space of an autoencoder. Recently, several approaches have been proposed to reduce the number of sampling steps~\cite{ddim, dpm, salimans2022progressive}. Unfortunately, such approaches usually compromise performance, especially for the lower number of steps.

Typically, diffusion-based models yield the best results on image patches of similar sizes to those seen during training (\eg $64 \times 64$ for SD \cite{ldm_cvpr22}). On the other hand, super-resolution applications require operating in high-resolution settings, drastically exacerbating the computational issues of diffusion-based models. For example, a SR model that aims for a magnification of $\times 4$ going from $256 \times 256$ to $1024 \times 1024$ requires dividing the input image into $16$ patches of $64 \times 64$ and running the model on each patch individually, making a large number of steps prohibitive for realistic use cases. Using state-of-the-art step-reduction strategy, such as more efficient samplers \cite{ddim,dpm,dpm++} can partially alleviate this issue but still falls widely short of practical needs. For example, going down to the target of 1 DDIM step results in a catastrophic drop in performance compared to a typical model that does 200 inference steps, as shown in Fig.~\ref{fig-step-vs-quality}.

One differentiating characteristic of the super-resolution task is that it is conditioned on the low-resolution (LR) input image to yield the target high-resolution (HR) image. Unlike the task of text-to-image generation, which relies on text conditioning, the LR image provides closer content to the target HR image, especially at lower scale factors. Therefore, conditioning the diffusion model on the LR image at low-scale factors makes the task inherently simpler for the diffusion model. In this paper, we take advantage of this peculiarity and introduce a novel training strategy dubbed scale distillation. While typical diffusion-based SR methods train the model for super-resolution by conditioning directly on the LR image at the target scale factor, we instead propose a progressive training approach, where we start by training a model for lower scale factors (\ie where the conditioning signal is closer to the target) and progressively increase to the target scale factor using the previously trained model as a teacher.

More specifically, instead of using the raw data to train a model for large scale factors, scale distillation obtains a rich and accurate supervisory signal from a teacher trained for a smaller scale factor. We first train a teacher that takes a less degraded image as input and, therefore, has an easier task to solve during training. Then, we train a model for a larger scale factor as a student while initializing it with the same weights as the teacher, which is now frozen. For a given time step during the training, we feed both teacher and student with the same noisy version of the HR image. However, we condition the teacher with the less degraded LR image (\ie using the same scale that was used during teacher training), while we condition the student on the target (more degraded) LR image. We use the teacher's prediction as a target to train the student for the larger scale factor. 

This training strategy has two direct advantages: i) Unlike typical training where the supervisory signal is somewhat ambiguous as the target is the same for all noise levels, our student receives its target from the teacher and is therefore adaptive to the noise level. ii) The target is more accurate, especially in terms of the finer detail, because the teacher takes a less degraded LR image as input. 

The proposed scale distillation approach allows the model to solve the SR task in fewer steps as we have simplified the task for the student. In fact, we show that models trained with our approach improve significantly when a few steps are used during the inference, \eg one step, see Fig.~\ref{fig:fid-vs-steps}. Therefore, a direct advantage of the proposed approach is that fine-tuning the decoder directly on top of the diffusion model becomes computationally tractable due to the single inference step required. Taking advantage of this fine-tuning, we show that You Only Need One Step (YONOS)-SR outperforms state-of-the-art diffusion-based SR methods that require a large number (\eg 200) of inference steps.

In summary, our contributions are threefold: \textbf{I)} We introduce scale distillation to train SD models with a more accurate and fine supervisory signal for image super-resolution tasks. \textbf{II)} We show that our proposed scale distillation strategy yields more efficient SD models that allow for directly fine-tuning the decoder on top of a frozen one-step diffusion model.~\textbf{III)} We show that combining scale distillation followed by decoder fine-tuning with the U-Net frozen yields state-of-the-art results on the super-resolution task, even at high magnification factors, while requiring only one step.

\section{Related work}\label{sec:related-work}
\paragraph{Real image super-resolution.}
Image super-resolution entails restoring a High Resolution (HR) image given its Low Resolution (LR) observation. Solving this task for real images is especially challenging given the dramatic differences in real-world image distributions \cite{survey,realsr_cvprw20,drealsr,dped}. These differences arise from varying image degradation processes, different imaging devices, and image signal processing methods, all of which are difficult to properly model and generalize. For this reason, real image super-resolution (or blind super-resolution) has received significant interest among the research community \cite{esrgan,realesrgan_iccvw21,realsr_cvprw20,drealsr,zhang2021designing,unsupervised-dasr,jie2022DASR,stablesr_arx23}. While some methods attempt to learn the degradation process \cite{yan2021, shunta2020, wan2020,fritch2019}, their success remains limited due to the lack of proper large scale training data \cite{survey}, even while using some unsupervised methods \cite{CycleGAN}. In contrast, more popular approaches tackle the problem by explicitly modeling the degradation pipeline to create synthetic LR-HR pairs to use for training \cite{shocher2018,liang2021,realesrgan_iccvw21, zhang2021designing}. Given, the wider success of the explicit degradation modeling approach, we elect to rely on the widely used RealESRGAN degradation pipeline \cite{realesrgan_iccvw21} in training our model.
\paragraph{Diffusion-based super-resolution.} Since the early SRCNN \cite{srcnn} method, many deep learning-based solutions for blind super-resolution have been proposed \cite{realsr_cvprw20,esrgan,zhang2021designing,CycleGAN,realesrgan_iccvw21,femasr_mm22,ldm_cvpr22,sr3,sr3+}. Early work proposed to take advantage of this space by using semantic segmentation probability maps for guiding SR \cite{wang2018}. Most recent methods aim at taking advantage of learned generative priors to simplify the inverse imaging problem of blind image super-resolution. Usually, methods following this paradigm \cite{esrgan,zhang2021designing,realesrgan_iccvw21} rely on GANs \cite{gan} and build on their generative priors. More recently, diffusion models showed remarkable generative capabilities yielding impressive results across a range of applications \cite{ldm_cvpr22,controlnet}. As such, in this paper, we follow several recent works \cite{sr3,sr3+,ldm_cvpr22,stablesr_arx23} and rely on diffusion-based generative models to tackle the super-resolution problem. While diffusion-based models achieve impressive results, their main shortcoming is the long inference time. Diffusion-based models require several inference steps through the model to yield a final output, thereby limiting their practical use. Therefore, in this paper, we tackle the important problem of speeding up the inference of diffusion-based super-resolution. 
\paragraph{Guided distillation.} Recognizing the inference speed shortcoming of diffusion models, several works have been proposed recently to address this issue \cite{salimans2022progressive,Meng_2023_CVPR,ddim,dpm,dpm++}. These methods can be categorized into two main approaches. One approach is to tackle this problem at inference time by either proposing more efficient samplers \cite{ddim,alexia2021} or relying on higher-order solvers \cite{dpm,dpm++}. More closely related to ours are methods that aim at directly training a diffusion model that can solve the generative problem at hand in fewer steps through \emph{temporal} distillation \cite{salimans2022progressive,Meng_2023_CVPR}. Our method tackles the problem at training time as well but we propose \emph{scale} distillation, where our main idea is to reduce the inference speed by progressively making the generative problem easier during training. Notably, our approach is orthogonal to temporal distillation and can be used in tandem with it. 
\section{YONOS-SR}
\label{sec:yonos}
In this section, we describe YONOS-SR, our diffusion-based model for image super-resolution. First, we present an overview of the image super-resolution framework with the latent diffusion models in Sec.~\ref{sec:sr_ldm}. We then discuss our proposed scale distillation method that allows us to improve the performance with fewer sampling steps, \eg 1-step, in Sec.~\ref{sec:scale-distill}. Finally, in Sec.~\ref{dec-ft}, we discuss how the 1-step diffusion model allows for fine-tuning a decoder directly on top of the diffusion model, with a frozen U-Net.

\subsection{Super-resolution with latent diffusion models}
\label{sec:sr_ldm}
Given a training set in the form of pairs of low and high-resolution images $(\mathbf{x}_h,\mathbf{x}_l)\sim p(\mathbf{x}_h,\mathbf{x}_l)$, the task of image super-resolution involves estimating the probability distribution of $p(\mathbf{x}_h|\mathbf{x}_l)$. The stable diffusion framework 
uses a probabilistic diffusion model applied on the latent space of a pre-trained and frozen autoendoer. Let us assume that $\mathbf{z}_h = \mathcal{E}(\mathbf{x}_h), \mathbf{z}_l = \mathcal{E}(\mathbf{x}_l)$ be the corresponding projection of a given low and high-resolution images $(\mathbf{x}_h,\mathbf{x}_l)$, where $\mathcal{E}$ is the pre-trained encoder. The forward process of the diffusion model, $q(\mathbf{z}|\mathbf{z}_h)$ is a Markovian Gaussian process defined as
\begin{align}
q(\mathbf{z}_t|\mathbf{z}_h) = \mathcal{N}(\mathbf{z}_t;\alpha_t \mathbf{z}_h, \sigma_t \mathbf{I}), \mathbf{z}=\{ \mathbf{z}_t | t \in [0,1] \}
\end{align}

\noindent where $\mathbf{z}$ denotes the latent variable of the diffusion model and $\alpha_t,\sigma_t$ define the noise schedule such that the log signal-to-noise ratio, $\lambda_t = \log [{\alpha_t^2/\sigma_t^2}]$, decreases with $t$ monotonically. During training, the model learns to reverse this diffusion process progressively, \ie estimate $p(\mathbf{z}_{t-1}|\mathbf{z}_t)$, to generate new data starting from noise.

The super-resolution objective function is derived by maximizing a variational lower bound of the data log-likelihood of $p(\mathbf{z}_h|\mathbf{z}_l)$ via approximating the backward denoising process of $p(\mathbf{z}_h|\mathbf{z}_t,\mathbf{z}_l)$. Note that, for super-resolution, the denoising process is conditioned on the low-resolution input, $\mathbf{z}_l$,  as well. This can be estimated by the function $\hat{\mathbf{z}}_\theta(\mathbf{z}_t,\mathbf{z}_l,\lambda_t)$ parametrized by a neural netowork. We can train this function via a weighted mean square error loss.

\begin{align}
\argmin_\theta~ \mathbb{E}_{\epsilon,t}[\omega(\lambda_t) ||\hat{\mathbf{z}}_\theta(\mathbf{z}_t,\mathbf{z}_l,\lambda_t) - \mathbf{z}_h||_2^2]
\label{eq:standad_loss}
\end{align}
over uniformly sampled times $t \in [0,1]$ and $\mathbf{z}_t = \alpha_t \mathbf{z}_h + \sigma_t \epsilon$, $\epsilon \sim \mathcal{N}(0,I)$. There are several choices of weighting function $\omega(\lambda_t)$. We use the so called  $\mathbf{v}$ parameterziation~\cite{salimans2022progressive}, $(1+\frac{\alpha_t^2}{\sigma_t^2})$, throughout this paper. 

The inference process from a trained model involves a series of sequential calls, \ie steps, of $\hat{\mathbf{z}}_\theta$,  starting from $\mathbf{z}_1 \sim \mathcal{N}(0,I)$,  where the quality of the generated image improves monotonically with the number of steps as shown in the qualitative examples of Fig~.\ref{fig-step-vs-quality} and quantitative results of Fig.~\ref{fig:fid-vs-steps}. Several methods have been proposed to reduce the number of required steps at inference time \cite{ddim,dpm,dpm++}. We use the widely used DDIM sampler in this paper \cite{ddim}, and unfortunately, we see that the performance drops drastically with an extremely low number of steps. In the following, we introduce scale distillation to alleviate this shortcoming.

\subsection{Scale distillation}\label{sec:scale-distill}
The complexity of the image super-resolution task increases with the scale factor (SF). For example, a model trained for a lower SF (\eg $\times 2$) takes as input a less degraded image compared to a larger SF (\eg $\times 4$). Therefore, a diffusion model trained for $\times 2$ magnification should require fewer inference steps to solve the HR image generation task compared to a model trained for the x4 scale factor. 

To alleviate the training complexity for larger scale factors, we build on this observation and propose a progressive scale distillation training strategy. In particular, we start by training a teacher for a lower SF that takes a less degraded image as input. We then use its prediction as a target to train the model for a higher factor as a student.

Let $N$ be the target SF of interest. Standard training involves making pairs of low and high-resolution images, where the low-resolution image is smaller than the HR image by a factor of $1/N$. The common approach for generating the training pairs is to gather a set of high-resolution images, perform synthetic degradation to obtain the corresponding low-resolution image and train a model that directly performs $\times N$ magnification \cite{realesrgan_iccvw21,stablesr_arx23,ldm_cvpr22} using eq.~\ref{eq:standad_loss}. Instead, we start with training a standard diffusion-based teacher that performs a lower SF, which takes a less degraded LR image, \eg $2/N$, as input and use its prediction to train the student.

More precisely, Let us assume $\hat{\mathbf{z}}_\phi, \hat{\mathbf{z}}_\theta$ be the teacher and student denonising models parameterized by $\phi, \theta$ respectively.
To train the student for a factor of $N$, we generate two degraded images for a given high-resolution image with factors $1/N,2/N$, with latent representations denoted by $\mathbf{z}_l,\mathbf{z}_l'$ respectively. That means $\mathbf{z}_l'$ is less degraded compared to $\mathbf{z}_l$. Similar to the standard diffusion model training, we sample random noise at $t$ and add it to the high-resolution image to obtain $\mathbf{z}_t$. The scale distillation loss will be:

\begin{align}
\argmin_\theta~ \mathbb{E}_{\epsilon,t} [\omega(\lambda_t) || \hat{\mathbf{z}}_\theta(\mathbf{z}_{t}, \mathbf{z}_l, \lambda_t) - \hat{\mathbf{z}}_\phi(\mathbf{z}_{t}, \mathbf{z}_l', \lambda_t) ||_2^2]
\label{eq:distillation_loss}
\end{align}

\begin{figure}[t]
\includegraphics[width=.98\columnwidth]{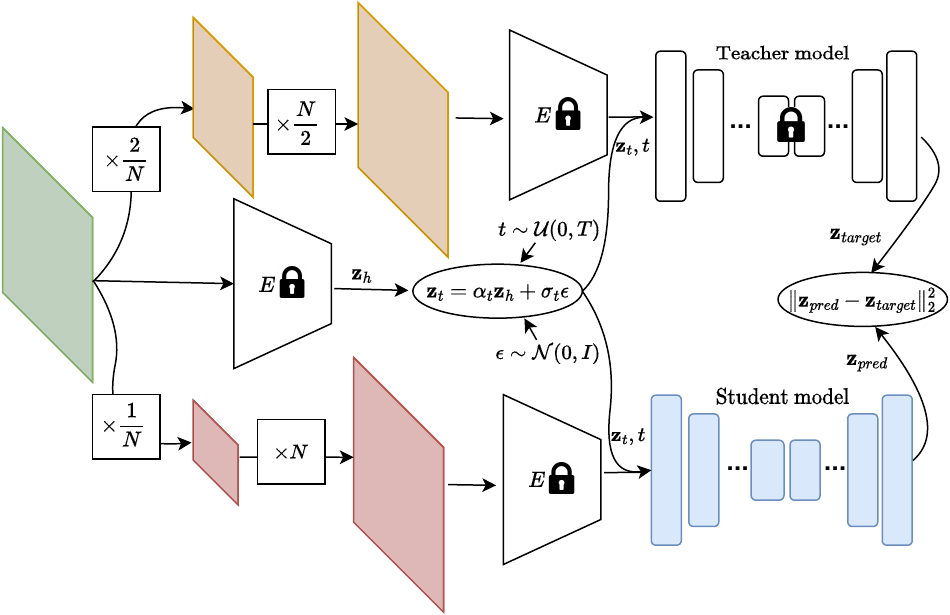}
\caption{Training pipeline of proposed scale distillation. For a given HR image (\eg size $512\times 512$) shown in green, we generate two degraded versions with factors of $2/N, 1/N$ (\eg sizes $256\times256$ and $128\times128$), shown in yellow and red respectively. Both degraded images are resized back via bicubic upsampling to $512 \times 512$ to be used as input to the encoder, which projects them to $4\times 64 \times 64$ tensors. The less and more degraded LR image is used as input to the teacher and student respectively via concatenation with the noisy version of the HR image, \ie $\mathbf{z}_t$. The teacher's output is used as the target for training the student. Note that the teacher is first trained independently for a smaller magnification scale and then frozen during student training. 
}
\label{fig:scale_dis}
\end{figure}

\noindent where the teacher is trained for $N/2$ magnification and frozen, and the student is initialized with the teacher's weights before the training. Note that we are using the latent diffusion framework that allows exactly the same architecture and input shapes for both the teacher and the student. Although the input low-resolution images for the student and teacher are of different sizes, they are both resized to a fixed size and fed to the encoder, which projects them to a tensor with a fixed size of $4\times 64 \times 64$. Fig.~\ref{fig:scale_dis} and Alg. ~\ref{alg:sd} illustrate the proposed scale distillation  process in detail. 

\begin{figure}[t]
  \begin{minipage}{0.235\textwidth}
      \begin{figure}[H]
          \centering
          \includegraphics[width=\linewidth]{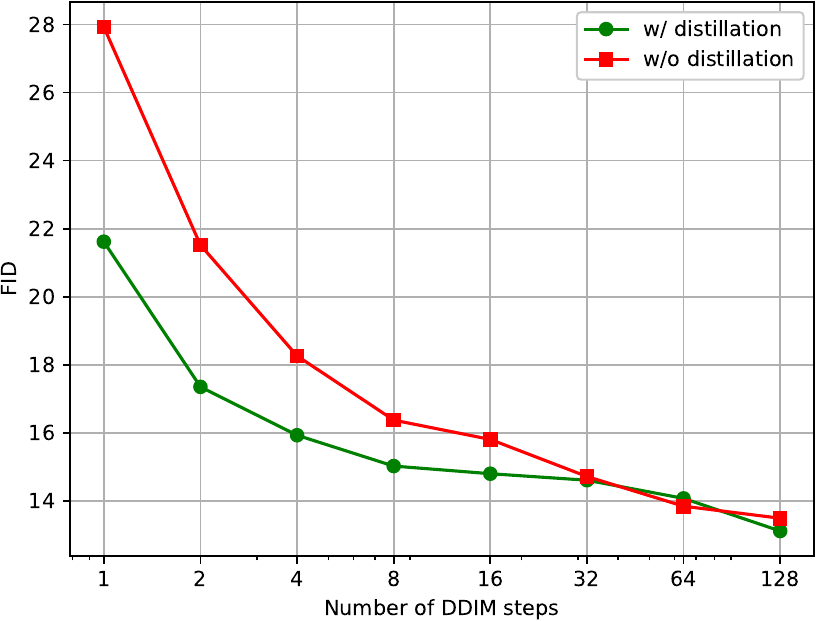}
          $\times 4$
      \end{figure}
  \end{minipage}
  \begin{minipage}{0.235\textwidth}
      \begin{figure}[H]
          \centering
          \includegraphics[width=\linewidth]{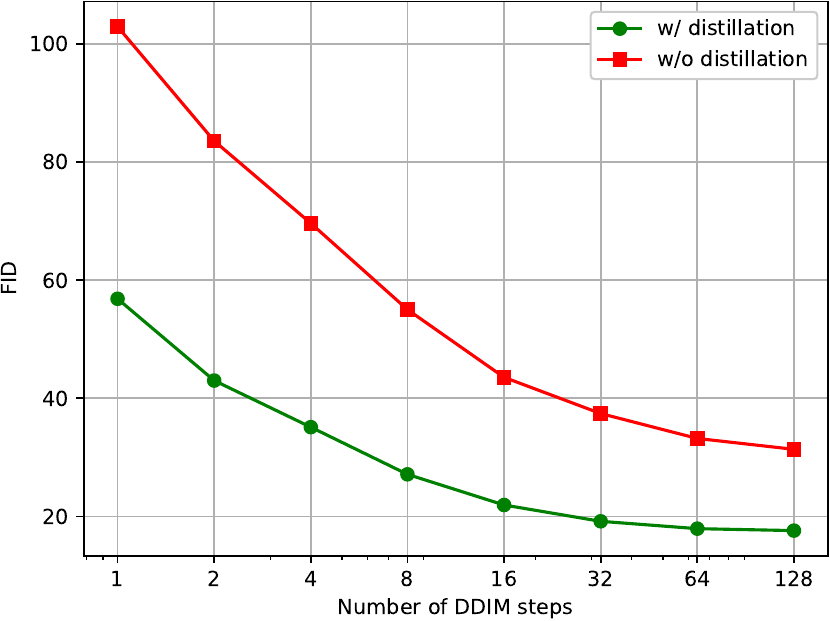}
          $\times 8$
      \end{figure}
  \end{minipage}
     \caption{FID vs. number of DDIM steps on the DIV2K validation set obtained through bicubic degradation for $\times 4$ and $\times 8$ magnifications.  We use $\times 2 \rightarrow \times 4$ scale distillation for $\times 4$ and $\times 2 \rightarrow \times 4 \rightarrow \times 8$ for $\times 8$ magnification,  and compare with the standard training directly for $\times 4$ and $\times 8$ respectively. All results are obtained using the original SD decoder. The model trained with scale distillation outperforms the standard training with large margin when using fewer steps for $\times 4$. The gap between scale distillation and the standard training is significantly higher for  $\times 8$ and remains steady for large numbers steps as well.}
   \label{fig:fid-vs-steps}
\end{figure}

\begin{algorithm}
\begin{algorithmic}
\State Input: dataset $\mathcal{D}$ 
\State Input: noise schedule $\alpha_t, \sigma_t, \lambda_t$ \Comment{for $t \in [0,1]$}
\State Input: scale factors $S$ \Comment{e.g. \{2,4,8\}} 
\State Input: initialization $\theta, \phi$
\Comment{from text-to-image}
\For{$i \in [0,\hdots,|S|]$}
    \State $s \gets S[i]$
    \While{not converged} \Comment{student training}
        \State $t \sim \mathcal{U}[0,1]$
        \State $\epsilon \sim \mathcal{N}(0,I)$
        \State $\mathbf{x}_h \sim \mathcal{D}$ 
        \State $\mathbf{x}_l \gets \Call{Degrade}{\mathbf{x}_h,s}$ 
        \State $\mathbf{z}_h \gets \mathcal{E}(\mathbf{x}_h)$
        \State $\mathbf{z}_l \gets \mathcal{E}(\Call{ResizeLike}{\mathbf{x}_l,\mathbf{x}_h})$
        \State $\mathbf{z}_t \gets \alpha_t \mathbf{z}_h + \sigma_t \epsilon$
        \If{$i > 0$}
            \State  \Comment{Obtain the target from the previous scale~~~}
            \State $s^\prime \gets S[i-1]$
            \State $\mathbf{x}^\prime_l \gets \Call{Degrade}{\mathbf{x}_h,s^\prime}$
            \State $\mathbf{z}^\prime_l \gets \mathcal{E}(\Call{ResizeLike}{\mathbf{x}^\prime_l,\mathbf{x}_h})$
            \State $\tilde{\mathbf{z}} \gets \hat{\mathbf{z}}_{\phi}(\mathbf{z}_t,\mathbf{z}^\prime_l,\lambda_t)$ 
        \Else
            \State  \Comment{Raw data as a target for the first teacher~~~~~}
            \State $\tilde{\mathbf{z}} \gets \mathbf{z}_{h}$
        \EndIf  
        \State $\mathcal{L}_\theta \gets \omega(\lambda_t)\|\hat{\mathbf{z}}_{\theta}(\mathbf{z}_t,\mathbf{z}_l,\lambda_t) - \tilde{\mathbf{z}}\|_2^2$
        \State $\theta \gets \theta - \eta \nabla_\theta \mathcal{L}_\theta$
    \EndWhile
    \State $\phi \gets \theta$
\EndFor
\end{algorithmic}
\caption{Scale Distillation Training. Given a set of scale factors, \eg. $\{2,4,8\}$, we start by training a student for the first scale  using the raw data(line~23) initialized with the text-to-image weights(line~4). We then use the trained student as a teacher to train the next distillation iteration for a higher magnification(line~20).  \textsc{DEGRADE} function degrades a given HR image with the given scale factor.  \textsc{RESIZE}\textsc{LIKE} function resizes a given LR image to the same size as the given HR image using the bicubic method.}
\label{alg:sd}
\end{algorithm}

The idea of scale distillation is in line with that of progressive temporal distillation~\cite{salimans2022progressive}. While a standard denoising model would only use the final image as the target irrespective of the sampled time step t (see Eq.~\ref{eq:standad_loss}), both scale and progressive temporal distillation rely on the teacher to provide a supervisory signal specific for step t (see Eq.~\ref{eq:distillation_loss}). In this way, the supervisory signal is attuned to the specific denoising step, providing stable and consistent supervision at every denoising step. 
Fig.~\ref{fig:fid-vs-steps} provides empirical support for our hypothesis. We observe a significant gap between the distilled model from $\times2$ to $\times4$ compared to the model that is directly trained for $\times4$ when evaluated with few inference steps. The gap shrinks as the number of steps increases and the quality starts saturating.

Similar to the temporal progressive distillation \cite{salimans2022progressive}, the proposed scale distillation process can be applied iteratively with higher scale factors at each training step. The first student is initialized from scratch and trained on the raw data, similar to the standard training. Consequently, this student becomes the new teacher for training the next scale factor. In this paper, we consider three distillation steps up to the scale factor of $\times8$ starting from $\times2$, \ie $\times 2\rightarrow \times 4 \rightarrow \times 8$. As it is shown in Fig.~\ref{fig:fid-vs-steps}, scale distillation is significantly more effective for $\times 8$ magnification where the LR image is of lower quality.

\subsection{Decoder fine-tuning}\label{dec-ft}

While scale distillation improves the one-step inference noticeably, there is still a gap between the one-step model and the saturated performance with a larger number of steps, see Fig.~\ref{fig:fid-vs-steps}. To fill this gap, we propose to fine-tune the decoder on top of the frozen one-step diffusion model resulting from scale distillation. That is, after training the diffusion model, we freeze the U-Net, apply one DDIM step for a given LR image, and use it as input to fine-tune the decoder for the SR task. We use the original loss that has been used for training the autoencoder \cite{ldm_cvpr22}. Importantly, this fine-tuning strategy with the U-Net in place is only possible with a diffusion model that can work properly with one step as enabled by our scale distillation approach; see Table.~\ref{tab:decoder_ablation}. We empirically show that the combination of our scale distillation approach with decoder fine-tuning yields a one-step model that can readily compete with models requiring a large number of inference steps.

\section{Experiments}\label{sec:experiments}
In this section, we evaluate our YONOS-SR against other methods targeting real image super-resolution at the standard $\times 4$ scale factor in Sec.~\ref{sec:x4comaprisons} and demonstrate that our proposed scale distillation approach generalizes to higher scale factors of $\times 8$ in Sec.~\ref{sec:x8comparisons}. We then provide qualitative results for $\times 4$ and $\times 8$ in Sec.~\ref{sec:qualitative}.  Finally,  we perform ablation studies in Sec.~\ref{sec:ablation} to highlight the role of our main contributions.

\begin{table*}[!htbp]
\resizebox{.99\textwidth}{!}{
    \centering
    \begin{tabular}{c|c|ccccc|cc|c}
        \hline
         Datasets & Metrics & RealSR & BSRGAN & DASR & Real-ESRGAN+ & FeMaSR & LDM & StableSR & YONOS (ours) \\
         \hline
        \multirow{5}{*}{\makecell{DIV2K Valid \\ RealESRGAN degradations} }
         & LPIPS $\downarrow$ & 0.5276 & 0.3351 & 0.3543 & 0.3112  & 0.3199 & \textcolor{red}{0.2510}& 0.3114 & \textcolor{blue}{0.2620}\\
         & FID $\downarrow$ & 49.49  & 44.22 & 49.16 & 37.64 & 35.87 & 26.47 & \textcolor{red}{24.44} & \textcolor{blue}{26.14}\\
         & MUSIQ $\uparrow$ & 28.57 & 61.19 & 55.19 & 61.05 & 60.83 & 62.27 & \textcolor{blue}{65.92} & \textcolor{red}{68.35} \\
        & PSNR $\uparrow$ & \textcolor{blue}{24.62} & 24.58 & 24.47 & 24.28 & 23.06 & 23.32 & 23.26 & \textcolor{red}{24.88}\\
         & SSIM $\uparrow$ & 0.5970 & 0.6269 & 0.6304 & \textcolor{blue}{0.6372} & 0.5887  & 0.5762 &  0.5726 & \textcolor{red}{0.6381}\\
         \hline
         \multirow{2}{*}{\makecell{DIV2K Valid \\ bicubic degradations}} & LPIPS $\downarrow$ & - & 0.2364 & 0.1696 & 0.2284 & - & 0.2323 & 0.2580 & \textcolor{red}{0.1534}\\
         & PSNR $\uparrow$ & - & \textcolor{blue}{27.32} & \textcolor{red}{28.55} & 26.65 & - & 25.49 & 21.90 & 26.71\\
         \hline
         \hline
         - & \#~STEPS $\downarrow$ & - & - & - & - & - & 200 & 200 & \textbf{\textcolor{red}{1}} \\
         \hline
         
         \hline
    \end{tabular}
    }
    \caption{Comparison to baselines on synthetic datasets. Results highlighted in \textcolor{red}{Red} and \textcolor{blue}{Blue} correspond to best and second best results, resp. Cells with $-$ indicate that there were no previously reported results using the considered baseline and the corresponding metric.}
    \label{tab:synthetic}
\end{table*}

\begin{table*}[ht]
\resizebox{.99\textwidth}{!}{
    \centering
    \begin{tabular}{c|c|ccccc|cc|c}
        \hline
         Datasets & Metrics & RealSR & BSRGAN & DASR & Real-ESRGAN+ & FeMaSR & LDM & StableSR & YONOS (ours) \\
         \hline
        \multirow{2}{*}{RealSR} & LPIPS $\downarrow$ & 0.3570 & \textcolor{blue}{0.2656} & 0.3134 & 0.2709 & 0.2937 & 0.3159 & 0.3002 & \textcolor{red}{0.2511}\\
        & MUSIQ $\uparrow$ & 38.26 & 63.28 & 41.21 & 60.36 & 59.06 & 58.90 & \textcolor{blue}{65.88} & \textcolor{red}{69.20}\\
         \hline
         \multirow{2}{*}{DRealSR}  & LPIPS $\downarrow$ & 0.3938 & \textcolor{blue}{0.2858} & 0.3099 & \textcolor{red}{0.2818} &  0.3157 & 0.3379 & 0.3284 & 0.3156\\
         & MUSIQ $\uparrow$ & 26.93 & 57.16 & 42.41 & 54.26 & 53.71 & 53.72 & \textcolor{blue}{58.51} & \textcolor{red}{65.02}\\
         \hline
         \multirow{1}{*}{DPED-iphone} & MUSIQ $\uparrow$ & 45.60 & 45.89 & 32.68 & 42.42 & 49.95 & 44.23 & \textcolor{blue}{50.48} & \textcolor{red}{58.76}\\
         \hline 
         \hline 
         - & \#~STEPS $\downarrow$ & - & - & - & - & - & 200 & 200 & \textbf{\textcolor{red}{1}} \\
         \hline 
         
         \hline
    \end{tabular}
    }
    \caption{Comparison to baselines on real datasets. Results highlighted in \textcolor{red}{Red} and \textcolor{blue}{Blue} correspond to best and second best results, resp.}
    \label{tab:real}
\end{table*}

\subsection{Evaluation on real image super resolution}\label{sec:x4comaprisons}
We first evaluate the performance of our proposed YONOS-SR model in the standard real image super-resolution setting targeting $\times 4$ scale factor.
\paragraph{Datasets.} Following previous work (\eg \cite{stablesr_arx23,femasr_mm22, realesrgan_iccvw21,zhang2021designing}), we use DIV2K \cite{div}, DIV8K\cite{div8k}, Flickr2k \cite{flickr2k}, OST \cite{ost} and a subset of 10K images from FFHQ training set \cite{ffhq} to train our model. We adopt the Real-ESRGAN \cite{realesrgan_iccvw21} degradation pipeline to generate synthetic LR-HR pairs. 

We then evaluate our model on both synthetic and real datasets. Similar to \cite{stablesr_arx23}, we use 3K LR-HR ($128\rightarrow512$) pairs synthesized from the the DIV2K validation set using the Real-ESRGAN degradation pipeline as our synthetic dataset. We also report results on the standard DIV2K validation split with bicubic degradations for completeness. For the real dataset, we use $128\times128$ center crops from the RealSR \cite{realsr_cvprw20}, DRealSR \cite{drealsr} and DPED-iphone \cite{dped} datasets.

\paragraph{Evaluation metrics.} We evaluate using various perceptual and image quality metrics, including LPIPS\cite{lpips}, FID \cite{fid} (where applicable), as well as the no-reference image quality metric, MUSIQ \cite{musiq}. For the synthetic datasets, we also report PSNR and SSIM, for reference.

\paragraph{Baselines.} As the main contribution of our paper targets improving the inference process of diffusion-based super-resolution, our main points of comparison are diffusion-based SR models, including the recent StableSR model \cite{stablesr_arx23} and the original LDM model \cite{ldm_cvpr22}.

For completeness, we also include comparison to other non-diffusion-based baselines, including; RealSR \cite{realsr_cvprw20}, BSRGAN \cite{zhang2021designing}, RealESRGAN \cite{realesrgan_iccvw21}, DASR \cite{jie2022DASR} and FeMaSR \cite{femasr_mm22}.

\paragraph{Results.} Results summarized in Tab. \ref{tab:synthetic} and \ref{tab:real} show that YONOS-SR outperforms all other diffusion-based SR methods, while using \textbf{only one inference step}, whereas other alternatives use 200 inference steps. These results highlight the efficiency of YONOS-SR in reducing the number of steps to one without compromising performance but indeed improving it further. Also, our model outperforms all considered baselines in 5 out of 7 metrics on the synthetic data and 4 out of 5 metrics on the real datasets. 

\subsection{Generalization to higher scale factors}\label{sec:x8comparisons}
We now evaluate the generalization capability of our proposed scale distillation approach. To this end, we train our YONOS-SR model with one more iteration of scale distillation, thereby going from a model capable of handling $\times 4$ magnifications to $\times 8$ magnifications. We then fine-tune the decoder on top of the one-step  $\times 8$ diffusion model. To evaluate this model, we follow recent work \cite{p2l}, and evaluate on the same subset of ImageNet and FFHQ for $\times 8$ magnification, \ie $64\times 64 \rightarrow 512\times 512$. In particular, we select the same $1\text{k}$ subset of Imagenet test set by first ordering the 10k images by name and then selecting the 1k subset via interleaved sampling, \ie using images of index 0, 10, 20, etc. To obtain the LR-HR pairs, we use $\times 8$ average pooling degradations. In the case of FFHQ, we use the first $1\text{k}$ images of the validation set. We also evaluate using the same metrics and baselines reported in this recent work \cite{p2l}. 

The results summarized in Tab.~\ref{tab:x8} demonstrate that our proposed one-step method generalizes well to higher scale factors, where it is able to achieve good results in terms of FID and LPIPS scores, which are known to better align with human observation, especially at higher magnification factors \cite{sr3+}. Notably, 
unlike baselines, our model has not been trained on ImageNet data. We use only $10\text{k}$ images of FFHQ in our training set.

\begin{table}[t]
\centering
\resizebox{\linewidth}{!}{
\begin{tabular}{l c c c c  c c c  }

\hline
 & \multicolumn{3}{c}{Imagenet} & & \multicolumn{3}{c}{FFHQ} \\
 \cline{2-4}
 \cline{6-8}
& FID $\downarrow$  & LPIPS $\downarrow$ & PSNR $\uparrow$ & & FID $\downarrow$  & LPIPS $\downarrow$ & PSNR $\uparrow$ \\ 
\hline
LDPS &  61.09 & 0.475 & \textcolor{blue}{23.21} & &  36.81 & 0.292 & \textcolor{red}{28.78} \\ 
GML-DPS~\cite{GML-DPS}  & 60.36 & 0.456 & \textcolor{blue}{23.21} &  & 41.65 & 0.318 & 28.50 \\ 
PSLD~\cite{GML-DPS} &  60.81 & 0.471 & 23.17 & &  36.93 & 0.335 & 26.62 \\ 
LDIR~\cite{LDIR} &  63.46 & 0.480 & 22.23 &  & 36.04 & 0.345 & 25.79 \\ 
P2L \cite{p2l} & \textcolor{blue}{51.81} & \textcolor{blue}{0.386} & \textcolor{red}{23.38} & &  \textcolor{blue}{31.23} & \textcolor{blue}{0.290} & \textcolor{blue}{28.55} \\ 
\hline
YONOS (ours) & \textcolor{red}{34.59} & \textcolor{red}{0.241} & 22.80 &  & \textcolor{red}{21.41}  & \textcolor{red}{0.161}  & 26.08 \\ 
\hline
\end{tabular}
}
\caption{Comparison to baselines on ImageNet subset with x8 magnification factor. Results highlighted in \textcolor{red}{Red} and \textcolor{blue}{Blue} correspond to the best and second best results, resp. The results for other methods are taken from \cite{p2l}.}\label{tab:x8}
\end{table}
\subsection{Qualitative evaluation}
\label{sec:qualitative}
\begin{figure*}[ht]
\begin{minipage}{0.99\textwidth}

  \begin{minipage}{0.245\textwidth}
      \begin{figure}[H]
          \centering
          \includegraphics[width=\linewidth]{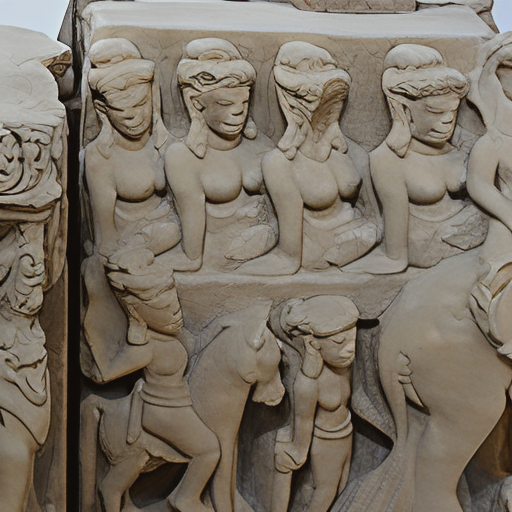}
          (a)
      \end{figure}
  \end{minipage}
  \begin{minipage}{0.245\textwidth}
      \begin{figure}[H]
          \centering
          \includegraphics[width=\linewidth]{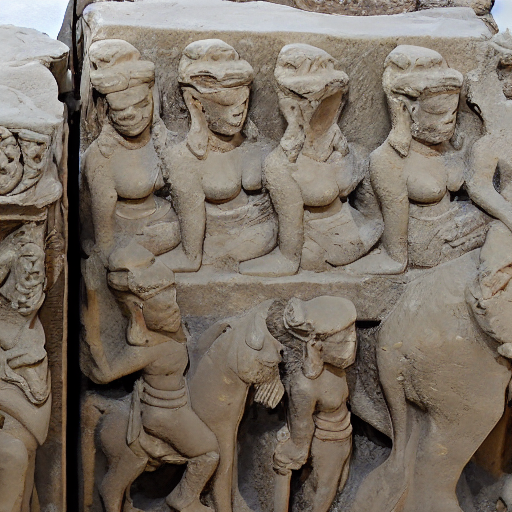}
          (b)
      \end{figure}
  \end{minipage}
   \begin{minipage}{0.245\textwidth}
      \begin{figure}[H]
          \centering
          \includegraphics[width=\linewidth]{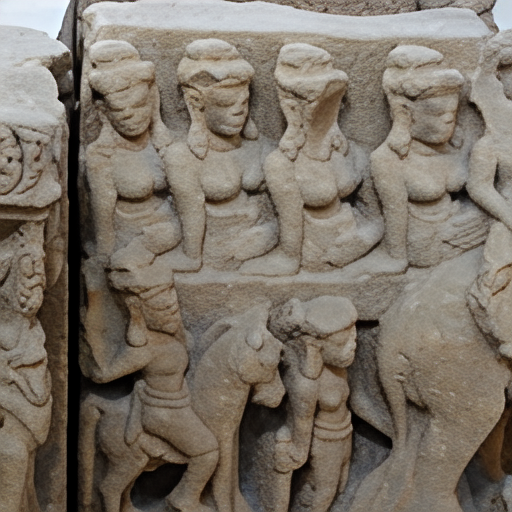}
          (c)
      \end{figure}
  \end{minipage}
   \begin{minipage}{0.245\textwidth}
      \begin{figure}[H]
          \centering
          \includegraphics[width=\linewidth]{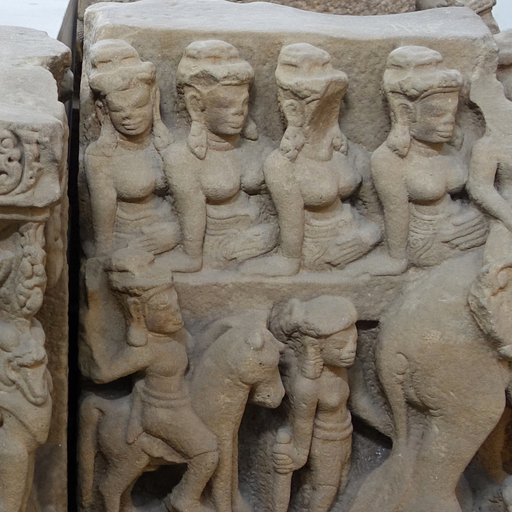}
          (d)
      \end{figure}
  \end{minipage}
  \\
  \begin{minipage}{0.245\textwidth}
      \begin{figure}[H]
          \centering
          \includegraphics[width=\linewidth]{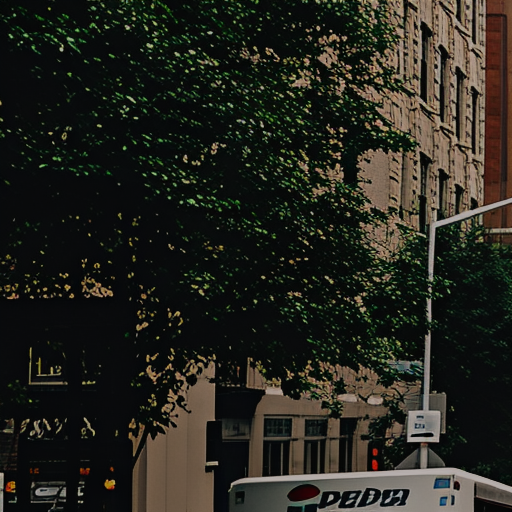}
          (a)
      \end{figure}
  \end{minipage}
  \begin{minipage}{0.245\textwidth}
      \begin{figure}[H]
          \centering
          \includegraphics[width=\linewidth]{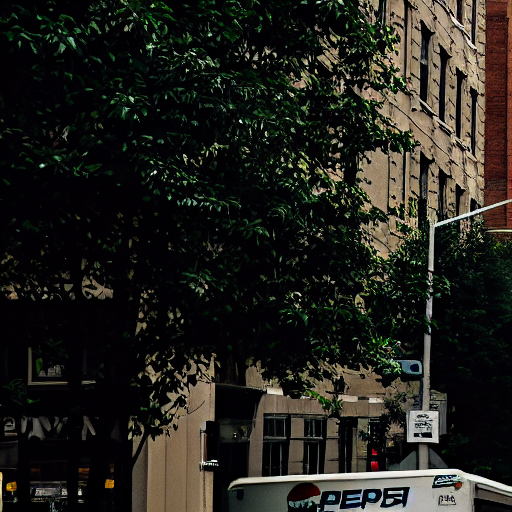}
          (b)
      \end{figure}
  \end{minipage}
   \begin{minipage}{0.245\textwidth}
      \begin{figure}[H]
          \centering
          \includegraphics[width=\linewidth]{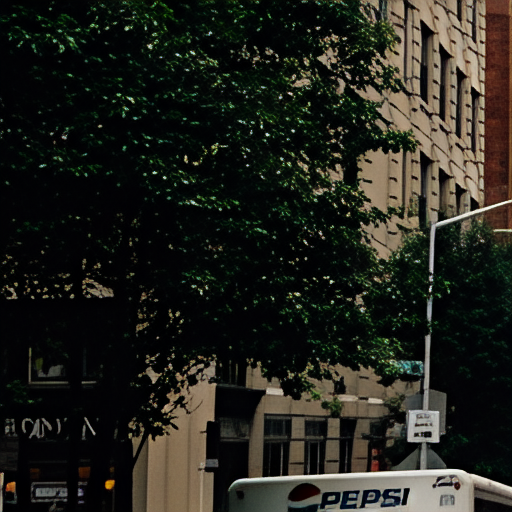}
          (c)
      \end{figure}
  \end{minipage}
   \begin{minipage}{0.245\textwidth}
      \begin{figure}[H]
          \centering
          \includegraphics[width=\linewidth]{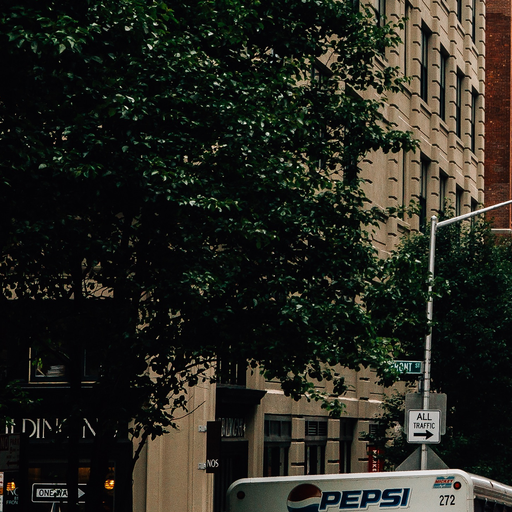}
          (d)
      \end{figure}
  \end{minipage}
      \\
  \begin{minipage}{0.245\textwidth}
      \begin{figure}[H]
          \centering
          \includegraphics[width=\linewidth]{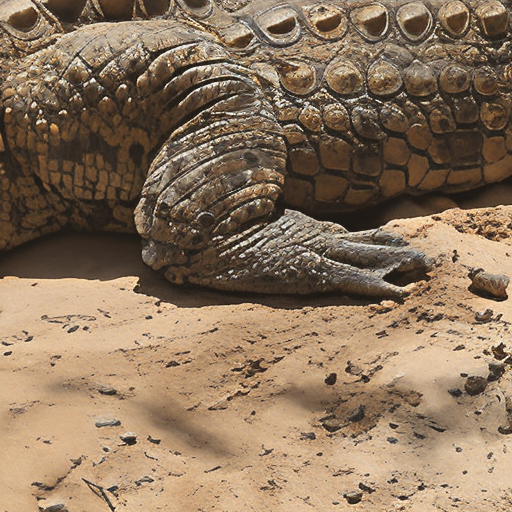}
          (a)
      \end{figure}
  \end{minipage}
  \begin{minipage}{0.245\textwidth}
      \begin{figure}[H]
          \centering
          \includegraphics[width=\linewidth]{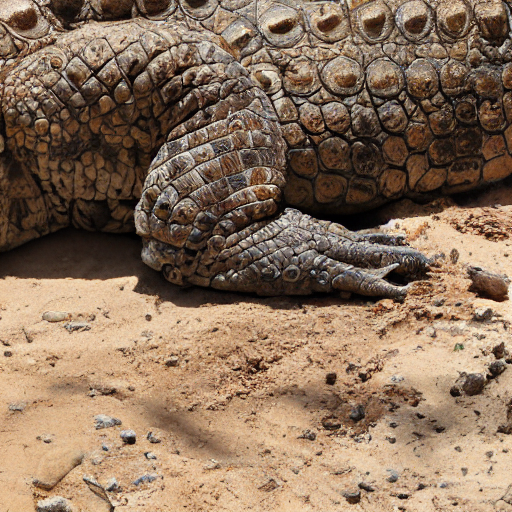}
          (b)
      \end{figure}
  \end{minipage}
   \begin{minipage}{0.245\textwidth}
      \begin{figure}[H]
          \centering
          \includegraphics[width=\linewidth]{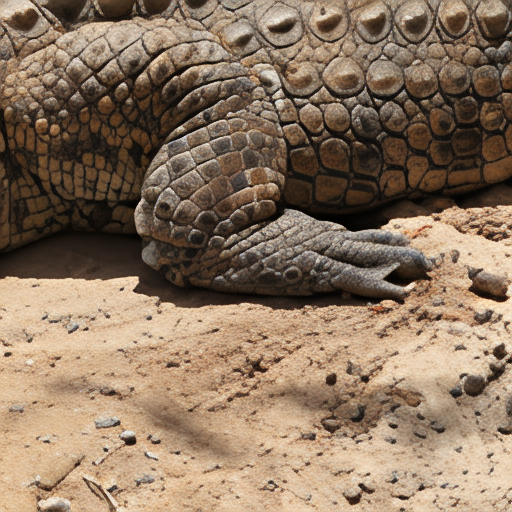}
          (c)
      \end{figure}
  \end{minipage}
   \begin{minipage}{0.245\textwidth}
      \begin{figure}[H]
          \centering
          \includegraphics[width=\linewidth]{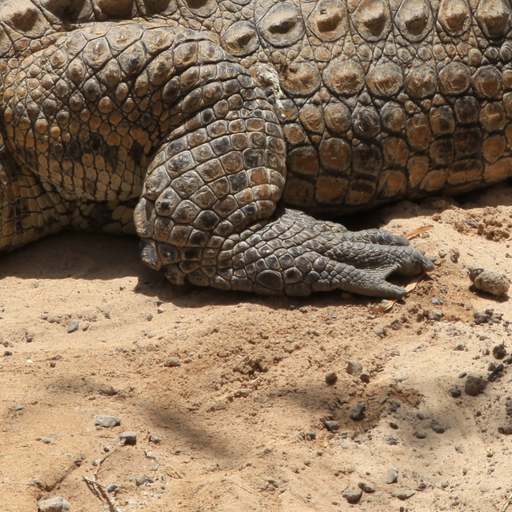}
          (d)
      \end{figure}
  \end{minipage}
\end{minipage}
\caption{Qualitative comparison on the validation set of DIV2K bicubic degradation dataset: (a) 200-step StableSR (b) 200-step standard SD-SR (c) \textbf{1-step} YONOS(ours) (d) the ground truth. SD-SR represents the standard Stable Diffusion-based SR model. 200-step StableSR and SD-SR tend to over-sharpen, adding artifacts that do not match the ground truth content. Our SR images match the most with the corresponding ground truth image; see the faces, Pepsi, and crocodile textures in the first, second, and third rows, respectively. The images are best seen in a display and zoomed in.}
\label{fig:qualititavie}
\end{figure*}

In addition to extensive quantitative evaluations, we qualitatively compare one-step YONOS-SR with 200-step StableSR and standard diffusion-based SR (SD-SR) in Fig.~\ref{fig:qualititavie}. Our method generates the closest SR images to the ground truth in terms of detailed textures while taking only \textbf{1-step} during the inference. These observations are in line with the numerical superiority of our method in the quantitative evaluations above. We perform two iterations of scale distillation $\times 2 \rightarrow \times 4$ and fine-tune the decoder on top of the 1 step model.

\begin{figure*}[ht]
\begin{minipage}{0.99\textwidth}
    
  \begin{minipage}{0.23\textwidth}
      \begin{figure}[H]
          \centering
          \includegraphics[width=\linewidth]{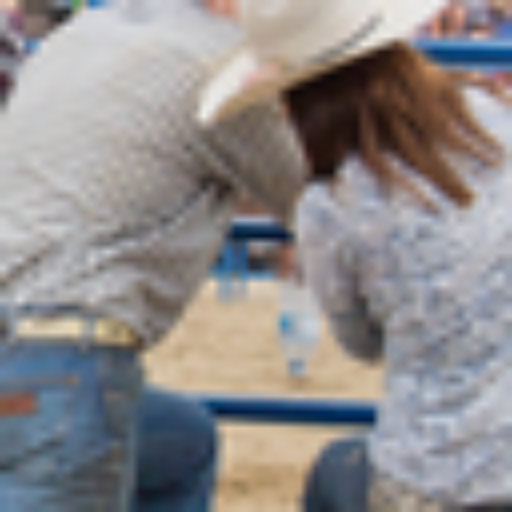}
          (LR)
          \end{figure}
    \end{minipage}
      \begin{minipage}{0.05\textwidth}
          \centering
          \rotatebox{90}{~~SD-SR }
          \rotatebox{90}{direct $\times 8$}
    \end{minipage}
    \begin{minipage}{0.23\textwidth}
     \begin{figure}[H]
          \centering
          \includegraphics[width=\linewidth]{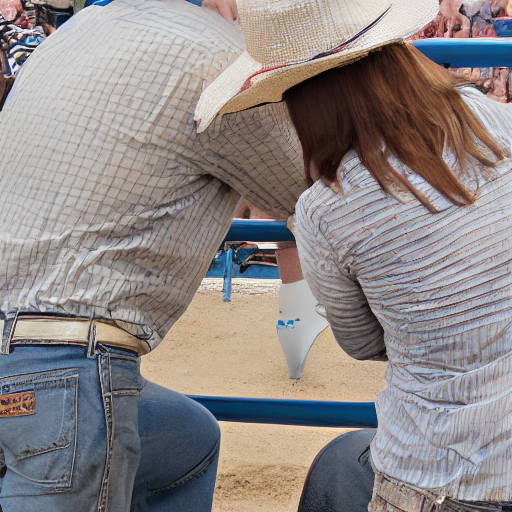}
          (64 steps)
      \end{figure}
  \end{minipage}
    \begin{minipage}{0.23\textwidth}
     \begin{figure}[H]
          \centering
          \includegraphics[width=\linewidth]{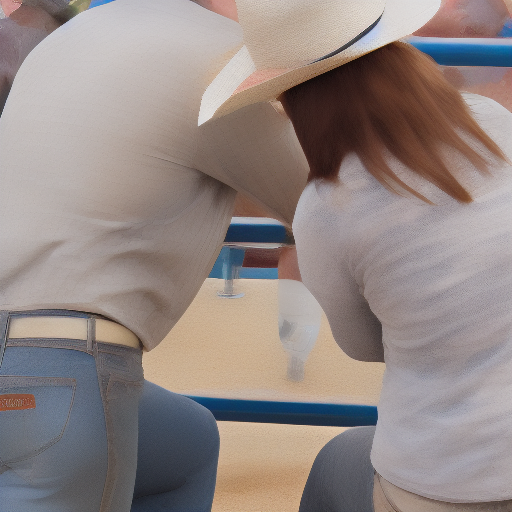}
          (4 steps)
      \end{figure}
  \end{minipage}
    \begin{minipage}{0.23\textwidth}
        \begin{figure}[H]
          \centering
          \includegraphics[width=\linewidth]{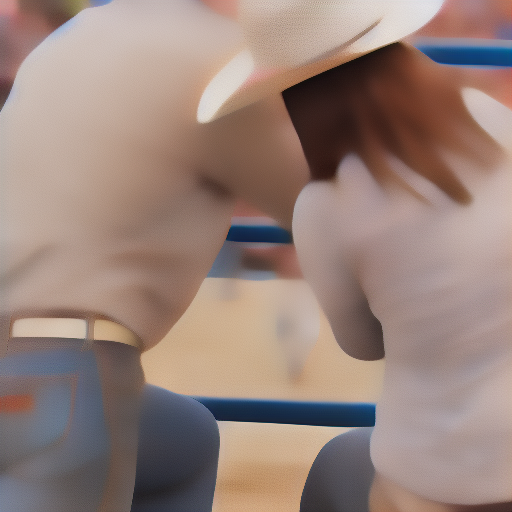}
          (1 step)
      \end{figure}
    \end{minipage}
   \begin{minipage}{0.23\textwidth}
     \begin{figure}[H]
          \centering
          \includegraphics[width=\linewidth]{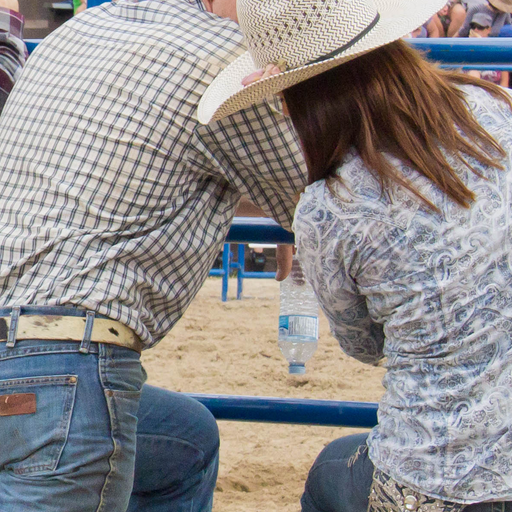}
          (HR)
      \end{figure}
  \end{minipage}
    \begin{minipage}{0.05\textwidth}
          \centering
          \rotatebox{90}{Scale distillation}
          \rotatebox{90}{$\times 2 \rightarrow \times 4 \rightarrow  \times 8$}
    \end{minipage}
      \begin{minipage}{0.23\textwidth}
     \begin{figure}[H]
          \centering
          \includegraphics[width=\linewidth]{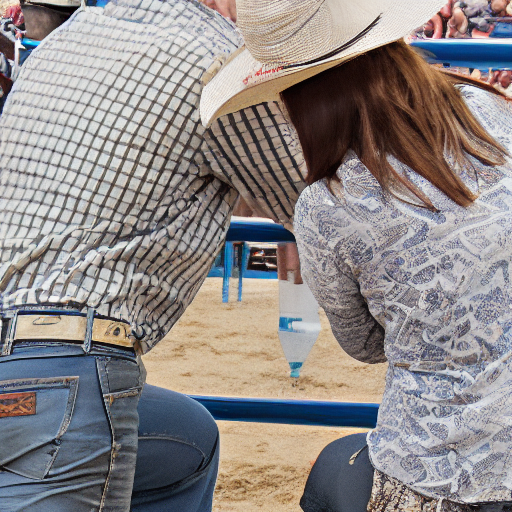}
          (64 steps)
      \end{figure}
  \end{minipage}
      \begin{minipage}{0.23\textwidth}
     \begin{figure}[H]
          \centering
          \includegraphics[width=\linewidth]{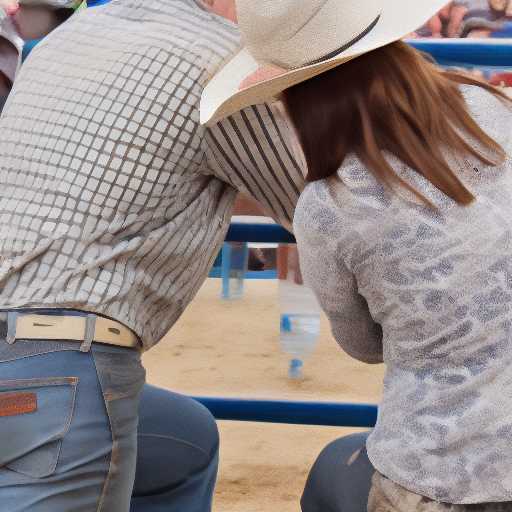}
          (4 steps)
      \end{figure}
  \end{minipage}
    \begin{minipage}{0.23\textwidth}
        \begin{figure}[H]
          \centering
          \includegraphics[width=\linewidth]{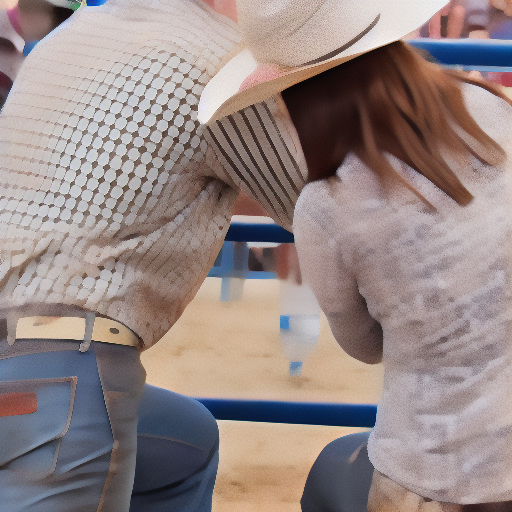}
          (1 step)
      \end{figure}
    \end{minipage}

    \begin{minipage}{0.23\textwidth}
      \begin{figure}[H]
          \centering
          \includegraphics[width=\linewidth]{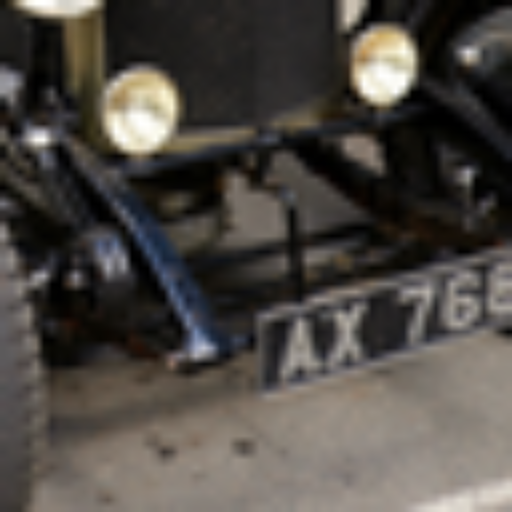}
          (LR)
          \end{figure}
    \end{minipage}
      \begin{minipage}{0.05\textwidth}
          \centering
          \rotatebox{90}{~~SD-SR }
          \rotatebox{90}{direct $\times 8$}
    \end{minipage}
    \begin{minipage}{0.23\textwidth}
     \begin{figure}[H]
          \centering
          \includegraphics[width=\linewidth]{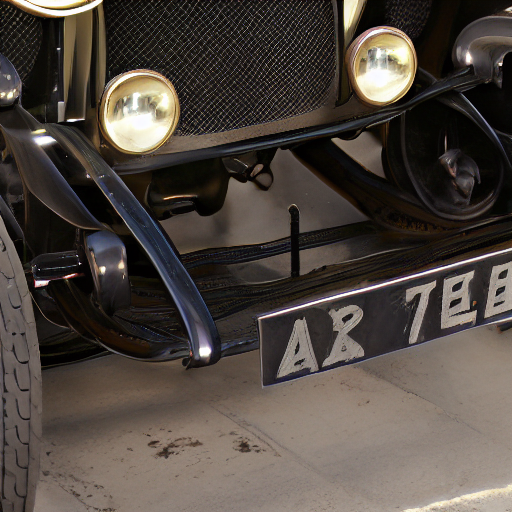}
          (64 steps)
      \end{figure}
  \end{minipage}
    \begin{minipage}{0.23\textwidth}
     \begin{figure}[H]
          \centering
          \includegraphics[width=\linewidth]{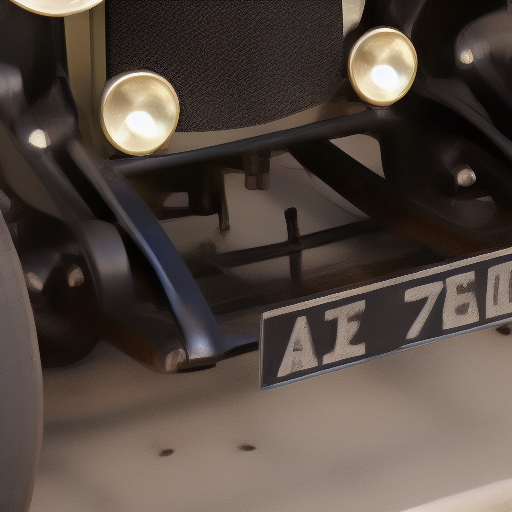}
          (4 steps)
      \end{figure}
  \end{minipage}
    \begin{minipage}{0.23\textwidth}
        \begin{figure}[H]
          \centering
          \includegraphics[width=\linewidth]{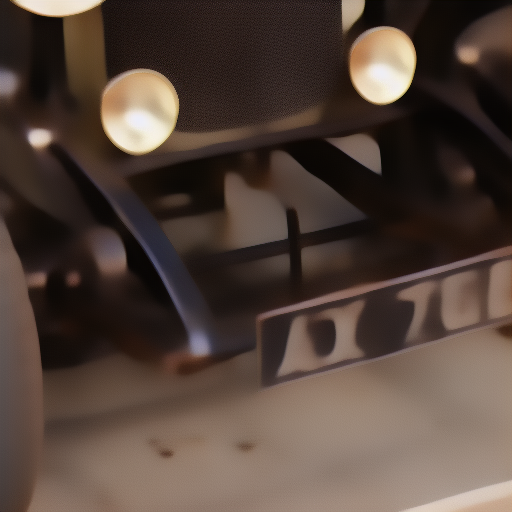}
          (1 step)
      \end{figure}
    \end{minipage}
   \begin{minipage}{0.23\textwidth}
     \begin{figure}[H]
          \centering
          \includegraphics[width=\linewidth]{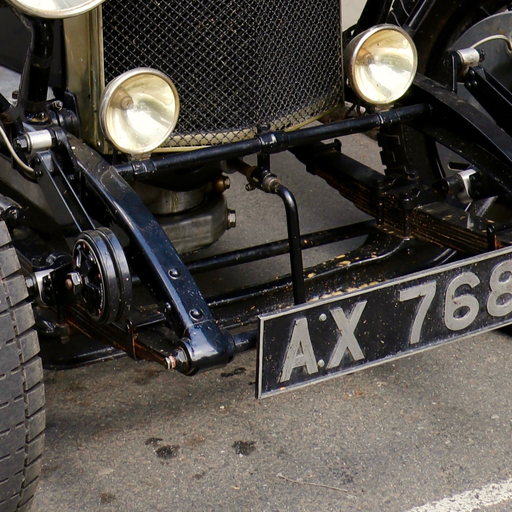}
          (HR)
      \end{figure}
  \end{minipage}
    \begin{minipage}{0.05\textwidth}
          \centering
          \rotatebox{90}{Scale distillation}
          \rotatebox{90}{$\times 2 \rightarrow \times 4 \rightarrow  \times 8$}
    \end{minipage}
      \begin{minipage}{0.23\textwidth}
     \begin{figure}[H]
          \centering
          \includegraphics[width=\linewidth]{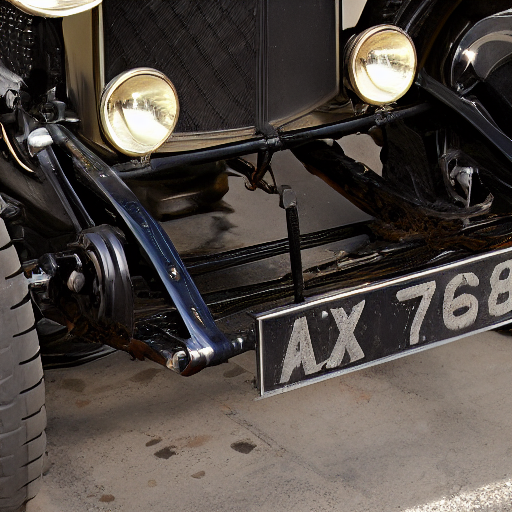}
          (64 steps)
      \end{figure}
  \end{minipage}
      \begin{minipage}{0.23\textwidth}
     \begin{figure}[H]
          \centering
          \includegraphics[width=\linewidth]{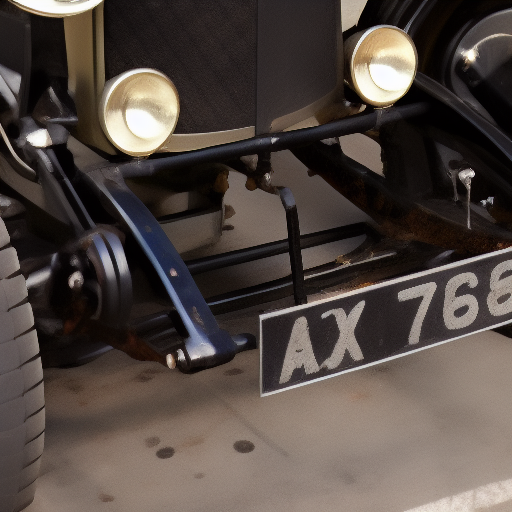}
          (4 steps)
      \end{figure}
  \end{minipage}
    \begin{minipage}{0.23\textwidth}
        \begin{figure}[H]
          \centering
          \includegraphics[width=\linewidth]{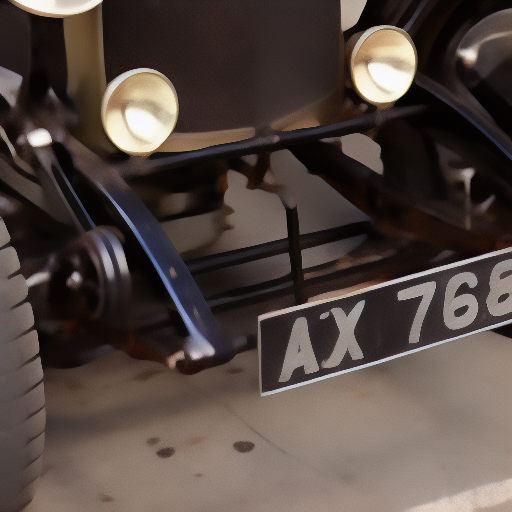}
          (1 step)
      \end{figure}
    \end{minipage}
  
\end{minipage}
\caption{~Qualitative comparison on the validation set of DIV2K bicubic degradation dataset for $\times 8$ magnification when the model is trained directly for $\times 8$ magnification without scale distillation (top row) and with three iterations of scale distillation $\times 2 \rightarrow \times 4  \rightarrow \times 8$ (bottom row). We show the input LR image,  the corresponding HR image,  and results with 1, 4, and 64 steps using the original decoder  for both models.  The model trained with scale distillation outperforms the standard training with high margins.  Specifically,  due to poor LR input, the standard training fails to recover the relevant content. The images are best seen in a display and zoomed in.}
\label{fig:qualititavie-x8}
\end{figure*}

As it is clearly demonstrated in Fig.~\ref{fig:fid-vs-steps}, scale distillation is significantly more effective for $\times 8$ compared to $\times 4$ magnification. As a qualitative support, we compare the model trained directly for $\times 8$ magnification without scale distillation  with three iterations of scale distillation $\times 2 \rightarrow \times 4  \rightarrow \times 8$ in Fig.~\ref{fig:qualititavie-x8}. Again, we use the validation set of DIV2K bicubic degradation dataset.  Following the numerical analyses in Fig.~\ref{fig:fid-vs-steps}, we observe that the model trained with scale distillation outperforms the standard training in terms of recovering the corresponding content and details. Note that, the problem of $\times 8$ magnification is of significantly higher complexity compared to  $\times 4$  due to poor LR input. Similar to Fig.~\ref{fig:fid-vs-steps}, we use the original decoder here to emphasize the impact of scale distillation.

\subsection{Ablation study}\label{sec:ablation}
In this section, we aim to study the impact of the various components introduced in our approach.
To this end, we use the standard DIV2K validation set with $
\times 4$ low-resolution images obtained through bicubic degradation~\cite{div}. We use the FID metric as it is a standard metric for assessing the quality of generative models. Our initial evaluation also revealed that the FID metric correlates the most with the human evaluation of the generated images. The validation set of the DIV2K dataset includes only 100 samples. To obtain more reliable FID scores, we extract $30$ random $128\times128$ patches and their corresponding $512\times512$ high-res counterparts from each image in the standard DIV2K bicubic validation set, resulting in a total of $3\text{k}$ LR-HR pairs. For completeness, we also report LPIPS, PSNR, and SSIM scores.

\paragraph{Impact of scale distillation.} We begin by evaluating the impact of our proposed scale distillation on speeding up inference time. To this end, we run two stable diffusions (SD) models trained for $\times 4$ super-resolution (SR), with various numbers of inference steps. The first model is a standard SD super-resolution model trained directly for target $\times 4$ super-resolution (\ie SD-SR), while the second model is trained with our proposed scale distillation from $\times 2$ magnification to $\times 4$. We use the same model, training set, and degradation pipeline in training both models. The only difference is the use of our scale distillation in the later model. Specifically, we start with training a teacher for $\times 2$ magnification using raw data as a denoising target. We use the $\times 2$ model as a frozen teacher and use its prediction to train a student for $\times 4$ magnification. The results summarized in Fig.~\ref{fig:fid-vs-steps} speaks decisively in favor of our scale distillation approach. We can see that for $\times 4$ magnification, the model trained without scale distillation needs at least \emph{twice} the number of inference steps that the model with scale distillation needs to reach a similar performance when the number of steps is smaller than $16$. Notably, we can see that our scale distillation model is performing especially well with as little as one inference step, where it outperforms the non-scale distilled baseline by at least 6 points. 

Scale distillation outperforms the standard training more significantly for $\times 8$ magnification where we perform three training iterations for scale distillation, \ie $\times 2 \rightarrow \times 4 \rightarrow \times 8$.  One reason for the larger gap for $\times 8$ magnification could be that the SR task is more ambiguous for $\times 8$ magnification due to lower quality input. As a result, the model benefits more from the more simplified supervisory signal obtained from scale distillation. Note that we use the original SD decoder model here only to analyze the impact of the scale distillation independently of decoder fine-tuning.

\paragraph{Impact of decoder fine-tuning.} One of the direct consequences of having a diffusion model that can yield good results in one denoising step is that it allows for decoder fine-tuning with the U-Net in place, as it will directly give a good starting point to the decoder. To validate the importance of the input given to the decoder prior to fine-tuning and, thereby, the importance of YONOS-SR, we experiment with the standard SD-SR model and our scale distillation model. In both cases, we freeze the U-Net and only allow the models to do 1 denoising step. We then feed their output to the decoder and fine-tune it following the same loss used in the original stable diffusion model \cite{ldm_cvpr22}. 

The results summarized in Tab.~\ref{tab:decoder_ablation} validate the importance of having a good initial input from the diffusion model prior to decoder fine-tuning. As we can see in the left chunk of Tab.~\ref{tab:decoder_ablation}, the model trained with scale distillation outperforms the standard training with a good margin when using the original decoder, indicating that the scale distillation results in a U-Net that provides a higher quality input for the decoder.  Moreover, as we can see in the right chunk of Tab.~\ref{tab:decoder_ablation}, fine-tuning the decoder on top of both 1-step models improves the performance. However, the model with scale distillation yields significantly better results than the standard SD-SR directly trained for the target magnification. The impact of scale distillation is more sensible for $\times 8$ magnification than $\times 4$, where FID improves from $41.54$ to $21.48$. Importantly, this fine-tuning strategy is not computationally feasible with diffusion models that require many denoising steps to give a reasonable starting point for the decoder. 

\begin{table}[h]
    \centering
    \begin{tabular}{cccc|cc|cc}
    \toprule
         & \multicolumn{2}{c}{Decoder}  && \multicolumn{2}{c|}{Original} & \multicolumn{2}{c}{Fine-tuned } \\
         \hline
          & \multicolumn{2}{c}{Scale distillation} & & \xmark & \checkmark   & \xmark & \checkmark \\
        \hline
         &  ~FID $\downarrow$ & & & 27.93 & \textbf{21.63} &  16.77  & \textbf{12.25} \\
          $\times 4$  & ~~~LPIPS $\downarrow$ & & & 0.227 & \textbf{0.210} & 0.162 & \textbf{0.145} \\
        & ~~~PSNR $\uparrow$ & & & 24.25 &  \textbf{25.06} & 24.21 & \textbf{25.19} \\ 
        & ~~~SSIM $\uparrow$ & & & 0.668 & \textbf{0.691} &  0.678 & \textbf{0.700} \\
        \hline
           &  ~FID $\downarrow$ & & & 102.92 & \textbf{56.84} &  41.54  & \textbf{21.48} \\
          $\times 8$ & ~~~LPIPS $\downarrow$ & & & 0.541 & \textbf{0.378} & 0.305 & \textbf{0.217} \\
          & ~~~PSNR $\uparrow$ & & & 21.08 &  \textbf{23.20} &  21.53 & \textbf{23.14} \\ 
         & ~~~SSIM $\uparrow$ & & & 0.541 & \textbf{0.610} &  0.528 & \textbf{0.610} \\
        \bottomrule
    \end{tabular}
    \caption{Role of the proposed scale distillation and decoder fine-tuning. All results reported here are obtained with 1 inference step.}
    \label{tab:decoder_ablation}
\end{table}

\section{Conclusion}
In summary, in this paper, we introduced the first \textbf{fast} stable diffusion-based super-resolution method. To achieve this, we introduced scale distillation, an approach that allows us to tackle the SR problem in as little as one step. Having a fast diffusion model allowed us to directly fine-tune the decoder, which we show yields state-of-the-art results, even at high magnification factors and only using a single step. We hope that the proposed distillation approach could be adapted for other inverse imaging problems (\eg image inpainting), which we believe is an interesting direction for future research.
{
    \small
    \bibliographystyle{ieeenat_fullname}
    \bibliography{main}
}
\end{document}